\documentclass[runningheads]{llncs}
\usepackage{graphicx}
\usepackage{tikz}
\usepackage{subfigure}
\usepackage{comment}
\usepackage{amsmath,amssymb}
\usepackage{color}
\usepackage{caption}
\usepackage{wrapfig}
\usepackage[title]{appendix}
\usepackage{enumerate}
\usepackage{enumitem}
\usepackage[accsupp]{axessibility}  

\usepackage[pagebackref,breaklinks,colorlinks,breaklinks=true]{hyperref}
\usepackage[capitalize]{cleveref}

\graphicspath{{./fig/}}

\newcommand{\captionMR}{Our 10-shot results on multiple artistic domains. From left to right: input face photos, Sketches, Caricature, Cartoon, Raphael, and Roy Lichtenstein results.}
\newcommand{\MRscale}{0.144}
\newcommand{\captionOneShot}{Our 1-shot results on multiple artistic domains, each with 1 training image shown in the top row.}
\newcommand{\ourGAN}{CtlGAN}

\begin{document}

\pagestyle{headings}
\mainmatter

\title{CtlGAN: Few-shot Artistic Portraits Generation with Contrastive Transfer Learning}


\titlerunning{CtlGAN}
\author{Yue Wang\inst{1} \and
Ran Yi\inst{1} \and
Luying Li\inst{1} \and
Ying Tai\inst{2} \and
Chengjie Wang\inst{2} \and
Lizhuang Ma \inst{1}}
\authorrunning{Y. Wang et al.}
\institute{Shanghai Jiao Tong University
\email{\{imwangyue,ranyi, liluying,ma-lz\}@sjtu.edu.cn} \and
Tencent Youtu Lab
\email{\{yingtai,jasoncjwang\}@tencent.com}}


\renewcommand\twocolumn[1][]{#1}%
\maketitle
\begin{center}
    \centering
    \captionsetup{type=figure}
    \includegraphics[width=\linewidth]{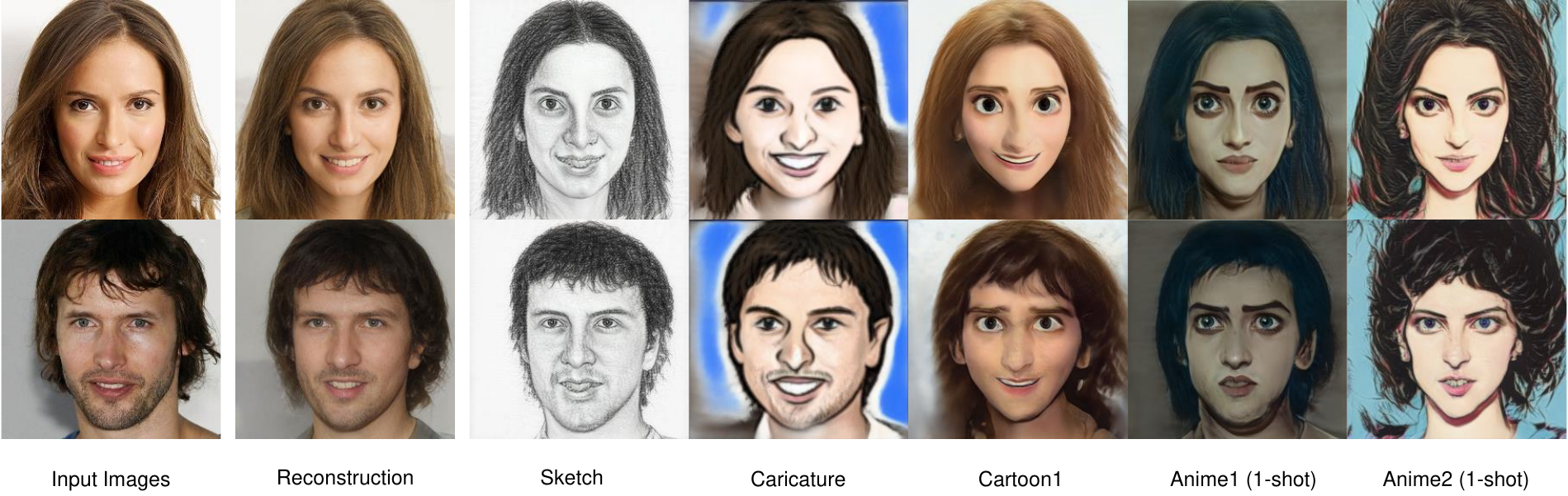}
    \vspace{-0.27in}
    \captionof{figure}{Our few-shot artistic portraits generation results on different artistic styles ({\it 10-shot or 1-shot}). 
    We eliminate  overfitting using a novel contrastive transfer learning strategy. With our style encoder, real face photos are embedded into the latent space shared by our decoders on different artistic domains.
    }
    \label{fig:teaser}
\end{center}

\vspace{-0.22in}
\begin{abstract}
Generating artistic portraits is a challenging problem in computer vision.
Existing portrait stylization models that generate good quality results are based on Image-to-Image Translation and require abundant data from both source and target domains.
However, without enough data, these methods would result in overfitting.
In this work, we propose \ourGAN, a new few-shot artistic portraits generation model with a novel contrastive transfer learning strategy.
We adapt a pretrained StyleGAN in the source domain to a target artistic domain with no more than 10 artistic faces.
To reduce overfitting to the few training examples, we introduce a novel Cross-Domain Triplet loss which explicitly encourages the target instances generated from different latent codes to be distinguishable.
We propose a new encoder which embeds real faces into $\mathcal{Z}+$ space and proposes a dual-path training strategy to better cope with the adapted decoder and eliminate the artifacts.
Extensive qualitative, quantitative comparisons and a user study show our method significantly outperforms state-of-the-arts under 10-shot and 1-shot settings and generates high quality artistic portraits.
The code will be made publicly available.

\end{abstract}

\section{Introduction}
\label{sec:intro}
Portrait art is a longstanding art form that captures human facial features in expressive art styles, such as painting, cartoon, sketch, and caricature.
However, even for professional artists, it takes hours to paint a good artistic portrait.
Developing computer programs to automatically generate artistic portraits can free artists from time-consuming and repeated works, and has the advantage of automatic portraits production with efficiency streamline.

With the development of machine learning, neural style transfer algorithms~\cite{GatysEB16,JohnsonAF16,huang2017arbitrary} are developed to transfer the style of a style exemplar to a content image.
However, these methods are unable to stylize portraits well since they tend to deform facial structures.
As Generative Adversarial Networks (GANs) gain success in various vision tasks, Image-to-Image Translation (IIT) methods leverage GANs to translate images from a source domain to a target domain by learning from paired~\cite{pix2pix,Wang0ZTKC18} or unpaired data~\cite{cyclegan,YiZTG17}. 
Based on this, some works~\cite{agilegan,jang2021stylecarigan,pinkney2020resolution,YiLLR19,zhu2021mind,chong2021jojogan} formulate the artistic portraits generation problem as the translation from real faces domain to artistic faces domain, and develop IIT algorithms to learn from a group of artistic faces.

However, these artistic portraits generation algorithms need abundant data, which is often difficult to acquire in real application scenarios.
For example, the Artistic-Faces Dataset~\cite{yaniv2019face} collects 160 artistic portraits of 16 different artists, only 10 for each artist, while existing methods often need at least 100 training images.
Although there are some research on few-shot Image-to-Image Translation ~\cite{liu2019few}, they mainly deal with translation between different object classes and few-shot generation for unseen classes, which is different from our problem.

We aim at learning a photo to artistic portrait translation by learning from a few artistic faces (e.g., no more than 10).
We observe that humans can learn artistic portraits of a certain style after seeing a small number of artistic samples, 
since they gain knowledge about faces in daily life, and apply it to portraits painting.
Similar to this, with transfer learning, machine applies knowledge gained in one problem to another related problem.
Although transfer learning has been explored in image generation with limited data~\cite{wang2018transferring,mo2020freeze,wang2020minegan,li2020ewc}, most methods still cannot generate good results when training examples are very few~\cite{ojha2021few-shot-gan}.
Recent research~\cite{ojha2021few-shot-gan} studies the image generation given only 10 training examples, by adapting a pretrained GAN to a target image domain via cross-domain correspondence. However, it didn't explicitly enforce the generations of different latent codes to be different, which leads to a certain degree of overfitting (Fig.~\ref{fig:compare_few-shot} middle).
Recently, some one-shot or text-guided methods~\cite{gal2021stylegan,zhu2021mind} were proposed leveraging the semantic power of CLIP~\cite{clip}, but these methods are worse in identity preservation.

In this paper, we propose \ourGAN, a novel few-shot artistic portraits generation model with {\it a contrastive transfer learning strategy}.
We adapt a pretrained StyleGAN2~\cite{Karras2019stylegan2} on real faces to a target artistic domain with no more than 10 artistic faces.
To prevent overfitting to the few training examples, we explicitly enforce the generations of different latent codes to be distinguishable with a new Cross-Domain Triplet loss.
To translate real faces into artistic portraits, we propose a new encoder to {\it invert} real faces into the StyleGAN2 latent space, which uses $\mathcal{Z}+$ latent space instead of $\mathcal{W}+$ and proposes a dual-path training strategy to cope with our decoder.
Our \ourGAN{} automatically generates high quality artistic portraits from real face photos under 10-shot or 1-shot settings (Figs.~\ref{fig:teaser},~\ref{fig:more_oneshot1}-\ref{fig:more_oneshot3},~\ref{fig:more_result0}-\ref{fig:more_result5}).

In summary, our main contributions are three-fold:
\begin{itemize}
\setlength{\itemsep}{1.5pt}
\item We propose \ourGAN{}, a new model for artistic portraits generation from real face photos under few-shot setting.
With no more than 10 training examples, our model generates high-quality artistic portraits for various artistic domains.

\item We present a novel contrastive transfer learning strategy that adapts a pretrained StyleGAN2 to a target artistic domain with Cross-Domain Triplet loss, and avoids overfitting to the few training samples.

\item We propose a novel style encoder which embeds real photos into $\mathcal{Z}+$ latent space and proposes a new dual-path training strategy to better cope with the adapted decoder and generate high-quality artistic portraits.
\end{itemize}

\section{Related Works}
\label{sec:related}

{\bf Generative Adversarial Networks.}
GANs~\cite{GoodfellowPMXWOCB14} achieve great success and are widely used in synthesizing images.
Conditional GANs~\cite{pix2pix} control the network outputs by conditional setting or inputs.
Recently, Karras et al. proposed StyleGAN series~\cite{stylegan,Karras2019stylegan2,Karras2020ada,karras2021alias} to improve the image synthesis quality and constructed a high quality face dataset named FFHQ.
Due to their high generation quality, StyleGAN series have achieved great success in many face generation tasks~\cite{tewari2020stylerig,abdal2021styleflow,YangRX021}.
We utilize StyleGAN2~\cite{Karras2019stylegan2} as the decoder and transfer a pretrained model on FFHQ to a target artistic portraits domain using no more than 10 examples with a novel contrastive transfer strategy.

{\bf GAN Inversion.}
GAN inversion is the process of embedding real images into the latent space of GANs.
In this paper, we focus on StyleGAN, where a $\mathcal{Z}$ space latent code is first translated into an intermediate $\mathcal{W}$ space by a mapping network, and then used to control the generator via AdaIN blocks~\cite{huang2017arbitrary} and output an image.
There are two main ways to realize GAN inversion: optimization based methods and learning based methods.
Optimization based methods try to optimize the latent code with some specialized loss functions.
Some works~\cite{Karras2019stylegan2,image2stylegan,image2stylegan++} find extending $\mathcal{W}$ space to $\mathcal{W}+$ space and directly optimizing $\mathcal{W}+$ space latent code can help improve the reconstruction quality.
Optimization based methods have higher quality in reconstructing images, while learning based methods process faster and use fewer computation resources.
Richardson et al. proposed pSp~\cite{richardson2021encoding}, an encoder based on a Feature Pyramid Network (FPN)~\cite{Lin2017FPN} to embed images to $\mathcal{W}+$ space.
Based on pSp architecture, e4e~\cite{e4e} proposed a new encoder for better StyleGAN-based image manipulation, and restyle~\cite{alaluf2021restyle} proposed iterative refinement for higher quality of reconstructed images.
Recently, Song et al. proposed a hierarchical Variational Autoencode (hVAE) in AgileGAN~\cite{agilegan} to embed face photos into latent codes that follow Gaussian distribution, and extend $\mathcal{Z}$ space to $\mathcal{Z}+$.
We adopt a similar $\mathcal{Z}+$ space setting, but instead of using a VAE, we propose a new encoder with dual-path training strategy, which eases the training and better copes with our adapted decoder.

\begin{figure}[t]
\subfigure[Comparison with \cite{ojha2021few-shot-gan} under 10-shot setting]{
	\begin{minipage}[t]{0.54\linewidth}
		\centering
        \includegraphics[width=0.9\linewidth]{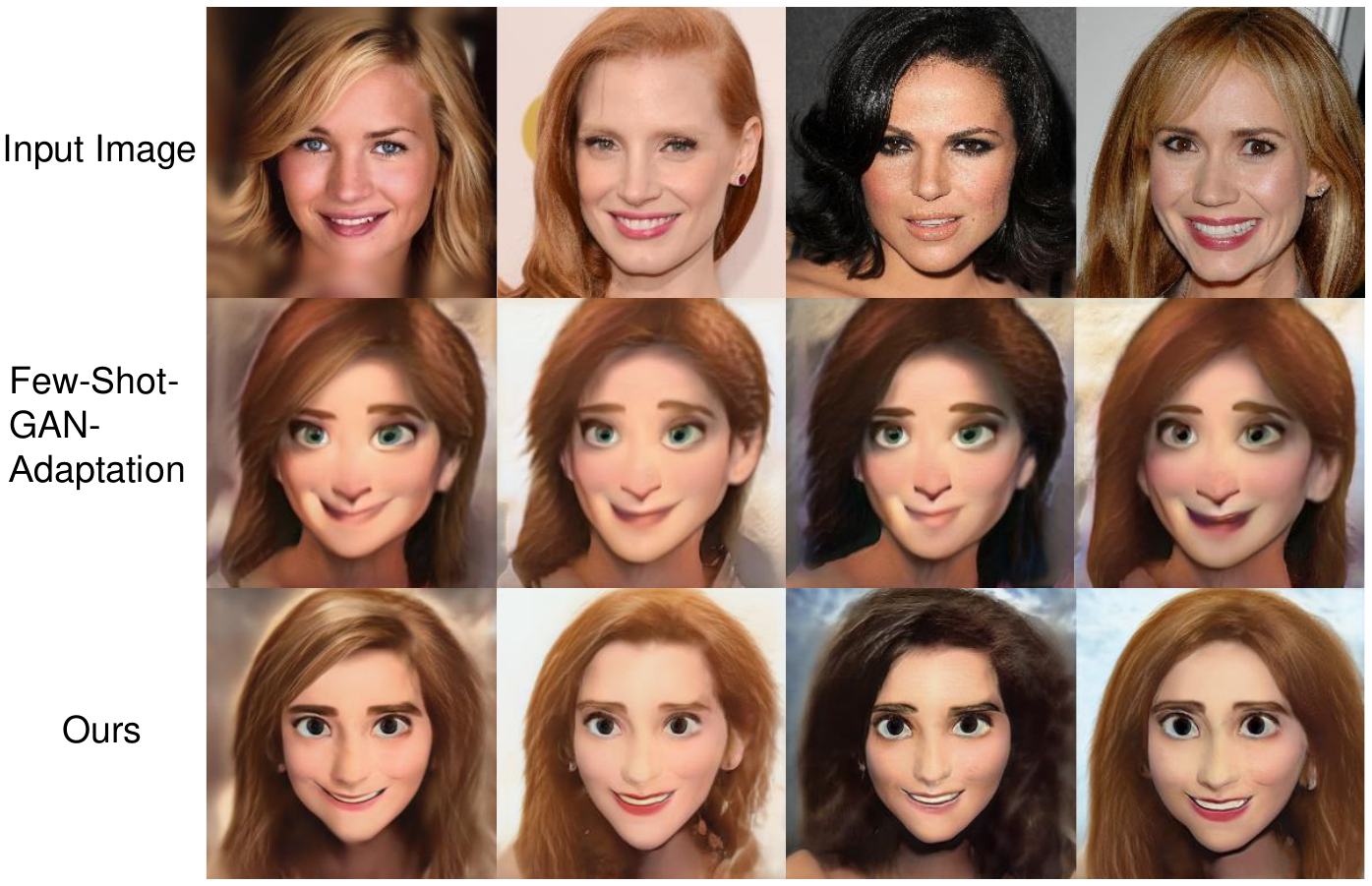}
        \vspace{-0.05in}
        \label{fig:compare_few-shot}
        \vspace{-0.15in}
	\end{minipage}
}
\subfigure[Encoder comparison ]{
	\begin{minipage}[t]{0.42\linewidth}
		\centering
        \includegraphics[width=0.9\linewidth]{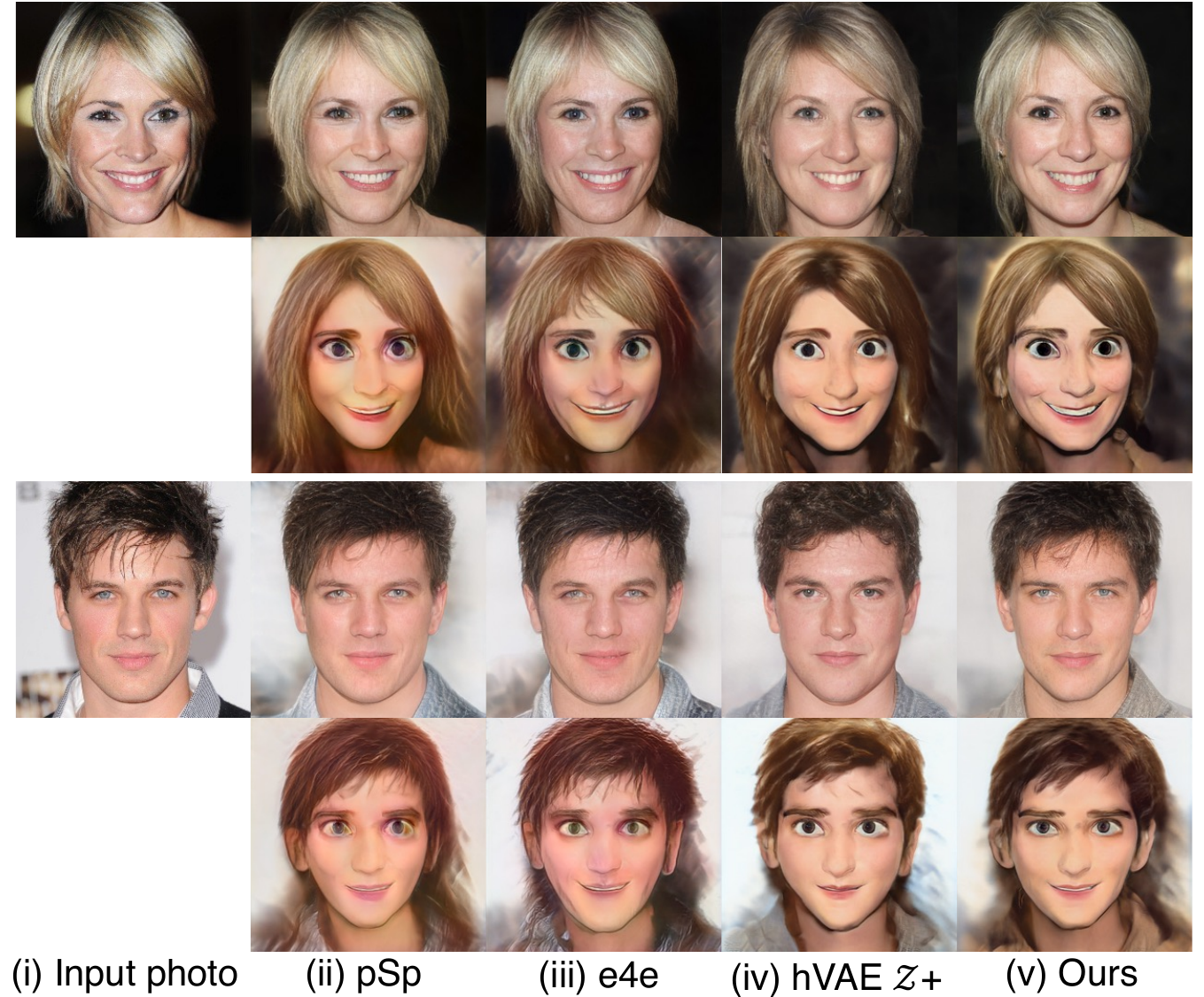}
        \vspace{-0.15in}
        \label{fig:compare_encoders}
	\end{minipage}
}
\vspace{-0.15in}
	\caption{(a) Few-Shot-GAN-Adaptation~\cite{ojha2021few-shot-gan} results show a certain degree of overfitting (similar faces in the middle row), while ours generates diverse results and well preserves the identity.
	(b) Results of combining different encoders~\cite{richardson2021encoding,e4e,agilegan} with our StyleGAN-based artistic decoder. For each input face, the top row shows reconstruction results, and the second row shows cartoonization results.}
\end{figure}

{\bf Transfer Learning for GANs.}
Transfer learning utilizes knowledge gained in solving one problem to solve related problems,
and helps the network training with limited data.
To help training GANs with limited data, some methods have been proposed to transfer GANs.
Transferring GANs (TGANs)~\cite{wang2018transferring} adapts a pretrained GAN model to a target domain by fine-tuning the original objective function.
BSA~\cite{noguchi2019image} only updates the scale and shift parameters in the generator during transfer.
FreezeD~\cite{mo2020freeze} freezes the lower layers of discriminator during adaptation.
MineGAN~\cite{wang2020minegan} proposes a miner network to find the knowledge that is most beneficial to a target domain from pretrained GANs.
EWC~\cite{li2020ewc} regularizes the weights changes during the adaptation, to best preserve the source ``information''.
However, these methods fail to generate good results when the training examples are very few~\cite{ojha2021few-shot-gan}. 
Few-Shot-GAN-Adaptation\cite{ojha2021few-shot-gan} transfers a pretrained GAN to a target domain with very few training samples, by preserving pairwise similarity before and after adaptation.
However, it is prone to overfitting (Fig.~\ref{fig:compare_few-shot} second row).
StyleGAN-NADA~\cite{gal2021stylegan} uses text to guide the domain adaptation by leveraging the semantic knowledge in the pretrained Contrastive-Language-Image-Pretraining (CLIP)~\cite{clip} model.
Mind-the-gap~\cite{zhu2021mind} also leverages the CLIP model, but uses a reference image (instead of text) to guide the domain adaptation. It builds upon StyleGAN-NADA and considers the semantic difference within domain.
These two methods utilize external knowledge from CLIP and achieve good adaptation results, but they are weaker in identity preservation.
JoJoGAN~\cite{chong2021jojogan} proposes a style mixing strategy to generate a large training set of paired face images from a single reference, and then finetunes StyleGAN using pixel loss, but sometimes suffers from artifacts.
In contrast, our \ourGAN{} transfers a pretrained GAN via contrastive transfer learning to address the overfitting and well preserves the identity.

{\bf Image-to-Image Translation.}
Image-to-Image Translation aims at translating images from a source domain to a target domain.
Early research works~\cite{pix2pix,Wang0ZTKC18} rely on paired data in source and target domain.
When paired data is not available, unpaired Image-to-Image Translation methods~\cite{cyclegan,YiZTG17,LiuBK17,HuangLBK18,ChoiCKH0C18,liu2019few} are proposed to utilize cycle consistency loss to learn from unpaired data.
Recently, Park et al. proposed a patch-based contrastive loss to constrain the patches of the translated images to match that of the source images at the sampled image locations.
Our work also utilizes the contrastive learning, but with a different format (adapt a source generator to a target generator) and a different purpose (prevent overfitting to the few training images).

{\bf StyleGAN-based Artistic Portrait Generation.}
Recently, some methods leverage StyleGAN for artistic portraits generation.
StyleCariGAN\cite{jang2021stylecarigan} and Toonify\cite{pinkney2020resolution} first train two StyleGAN networks on real face photos domain and artistic portraits domain, and swap some layers of the two StyleGANs to generate caricatures and cartoonized portraits. 
AgileGAN\cite{agilegan} transfers a pretrained StyleGAN on FFHQ to a stylistic domain and proposes a hierarchical VAE to embed real face photos to an extended $\mathcal{Z}$, i.e., $\mathcal{Z}+$ space, to translate real face photos into artistic styles.
However, the above methods need abundant data to train StyleGAN on artistic portraits domain or transfer a pretrained StyleGAN\footnote{StyleCariGAN uses 6,000+ caricatures, Toonify uses about 300 cartoon faces, and AgileGAN uses about 100 artistic faces.}. 
In comparison, our \ourGAN{} generates high quality results by learning from no more than 10 artistic examples.

\begin{figure*}[t]
  \centering
  \includegraphics[width=0.97\linewidth]{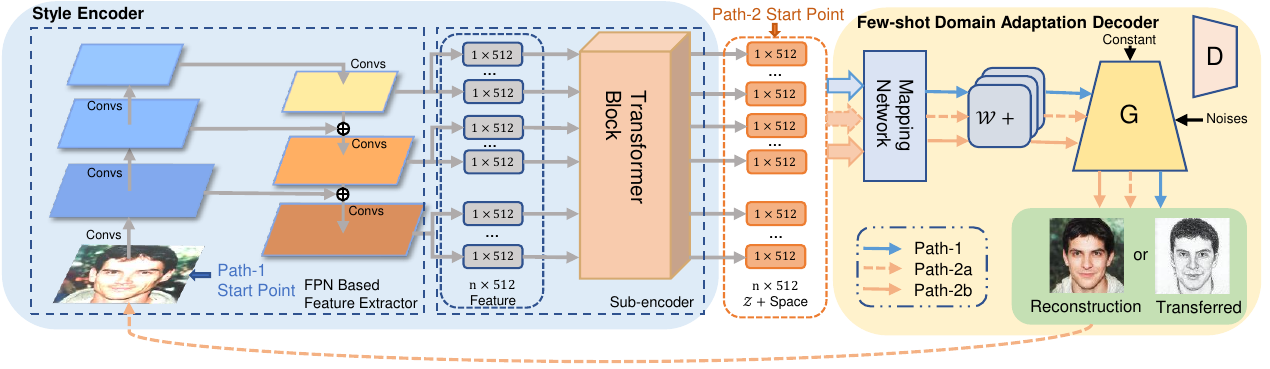}
  \vspace{-0.1in}
  \caption{Our \ourGAN{}
  contains two parts, including a Style Encoder, and a Domain Adaptation Decoder which is based on StyleGAN2~\cite{Karras2019stylegan2}.
  We design a dual path training for our encoder with path-1 shown in blue and path-2 (cycle path) shown in orange.
   }
  \label{fig:pipeline}
  \vspace{-0.15in}
\end{figure*}

\section{Method}
\label{sec:method}
Given a few (e.g., no more than 10) artistic examples, our task is to learn a  model which generates artistic portraits from real face photos. 
To solve this task, we design a novel \ourGAN{} with a contrastive transfer learning strategy and a style encoder.
Given a real face photo as input, our pipeline first encodes the face into a latent code, and then the decoder utilizes
the latent code to generate an artistic portrait.

We design a novel contrastive transfer learning strategy to train our decoder for artistic portraits generation. 
Noticed that StyleGAN series~\cite{stylegan,Karras2019stylegan2} have achieved great success in high quality face generation, we  leverage the facial knowledge in a pretrained StyleGAN2~\cite{Karras2019stylegan2} in real faces domain, and adapt the model to a target artistic domain.
To prevent overfitting to the few training samples, we propose a novel Cross-Domain Triplet loss, which explicitly enforces the target instances generated from different latent codes to be distinguishable.

\begin{figure*}[t]
  \centering
  \includegraphics[width=0.97\linewidth]{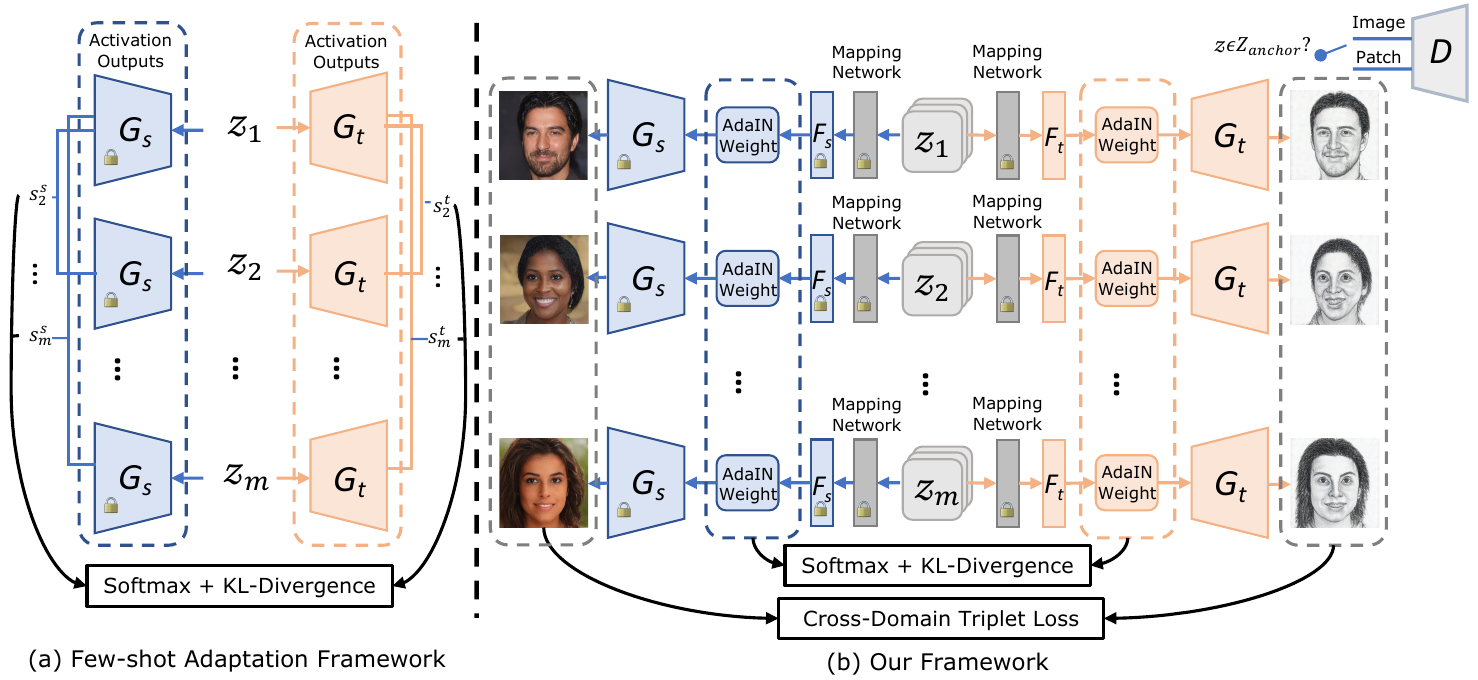}
   \vspace{-0.1in}
   \caption{(a) Few-Shot-GAN-Adaptation~\cite{ojha2021few-shot-gan} framework and (b) Our contrastive transfer learning framework. 
   Given a pretrained source generator and 10 target domain examples, Few-Shot-GAN-Adaptation adapts the model to the target domain by constraining the pairwise similarity before and after adaptation. While we explicitly enforce the target instances generated from two different latent codes to be different to prevent overfitting.
   }
   \label{fig:decoder_arch}
   \vspace{-0.15in}
\end{figure*}

In order to translate a real face photo into an artistic portrait while keeping the original identity, a decent encoder is needed to map the face photo into the latent space of StyleGAN.
Although there are some encoders~\cite{richardson2021encoding,e4e} to embed an image into the $\mathcal{W}+$ latent space (the extended $\mathcal{W}$ space) of StyleGAN, they can't well cope with our decoder and the results show obvious artifacts (Fig.~\ref{fig:compare_encoders}(ii-iii)). 
Recent research~\cite{agilegan} embeds real faces into a $\mathcal{Z}+$ space (extended $\mathcal{Z}$) to resolve the artifacts,
but this does not generate an accurate reconstruction and causes identity loss (Fig.~\ref{fig:compare_encoders}(iv)).
We follow the $\mathcal{Z}+$ space setting and propose a novel style encoder to better preserve the identity information, which consists of a feature extractor and a sub-encoder (Fig.~\ref{fig:compare_encoders}(v)). 
We also design a dual path training to constrain the encoder output close to Gaussian distribution.

As described above, 
our \ourGAN{} consists of two components: 
1) Few-shot Domain Adaptation Decoder (Sec.~\ref{sec:decoder}),  which is transferred from a pretrained StyleGAN to a target domain, and 2) Style Encoder (Sec.~\ref{sec:encoder}), which embeds real faces to $\mathcal{Z}+$ space latent codes~\cite{agilegan}.
The pipeline is shown in Fig.~\ref{fig:pipeline}.

\subsection{Few-shot Domain Adaptation Decoder}
\label{sec:decoder}
Given a pretrained StyleGAN2 model $\mathcal{G}_s$ trained on FFHQ dataset, which maps a $\mathcal{Z}$ space latent code to a realistic face image, we aim to transfer $\mathcal{G}_s$ to a target domain generator $\mathcal{G}_{t}$ for an artistic style, {\it using only a few examples} (e.g., 10).
Since adaption using very few examples easily leads to overfitting, to solve this problem, \cite{ojha2021few-shot-gan} proposes to preserve the relative distance before and after adaptation (Fig.~\ref{fig:decoder_arch}(a)).
However, its results still show similar facial features (Fig.~\ref{fig:compare_few-shot}), which lowers the identity similarity.

To prevent the generations from overfitting to the few training examples, we want target generations from two different latent codes $\mathcal{G}_{t}(z_i), \mathcal{G}_{t}(z_j)$ (pair1) to be different.
To better preserve identity, we want source and target generations from the same latent code $\mathcal{G}_{s}(z_i), \mathcal{G}_{t}(z_i)$ (pair2) to be similar in content (the same person with different styles). 
We propose a contrastive learning strategy (Fig.~\ref{fig:decoder_arch}(b)) to achieve the above properties as follows:

{\bf Cross-Domain Triplet loss.}
We propose to use triplet loss\cite{schroff2015facenet} to enforce the desired distance between pair1 and pair2.
The triplet is an image $x^a$ ({\it anchor})  with its {\it positive} example $x^p$ (same class) and its {\it negative} example $x^n$ (different class), and the triplet loss is calculated as:

\begin{equation}
  \begin{aligned}
  \mathcal{L}_{triplet} = \max(d(x^a, x^p) - d(x^a,x^n) + \alpha, 0),
  \label{eq:triplet_loss}
  \end{aligned}
\end{equation}
where $d$ measures the distance between two images, and $\alpha$ is a margin enforced between positive and
negative pairs.

\begin{wrapfigure}{r}{0.25\textwidth}
\vspace{-10pt}
\begin{center}
\includegraphics[width=0.25\textwidth]{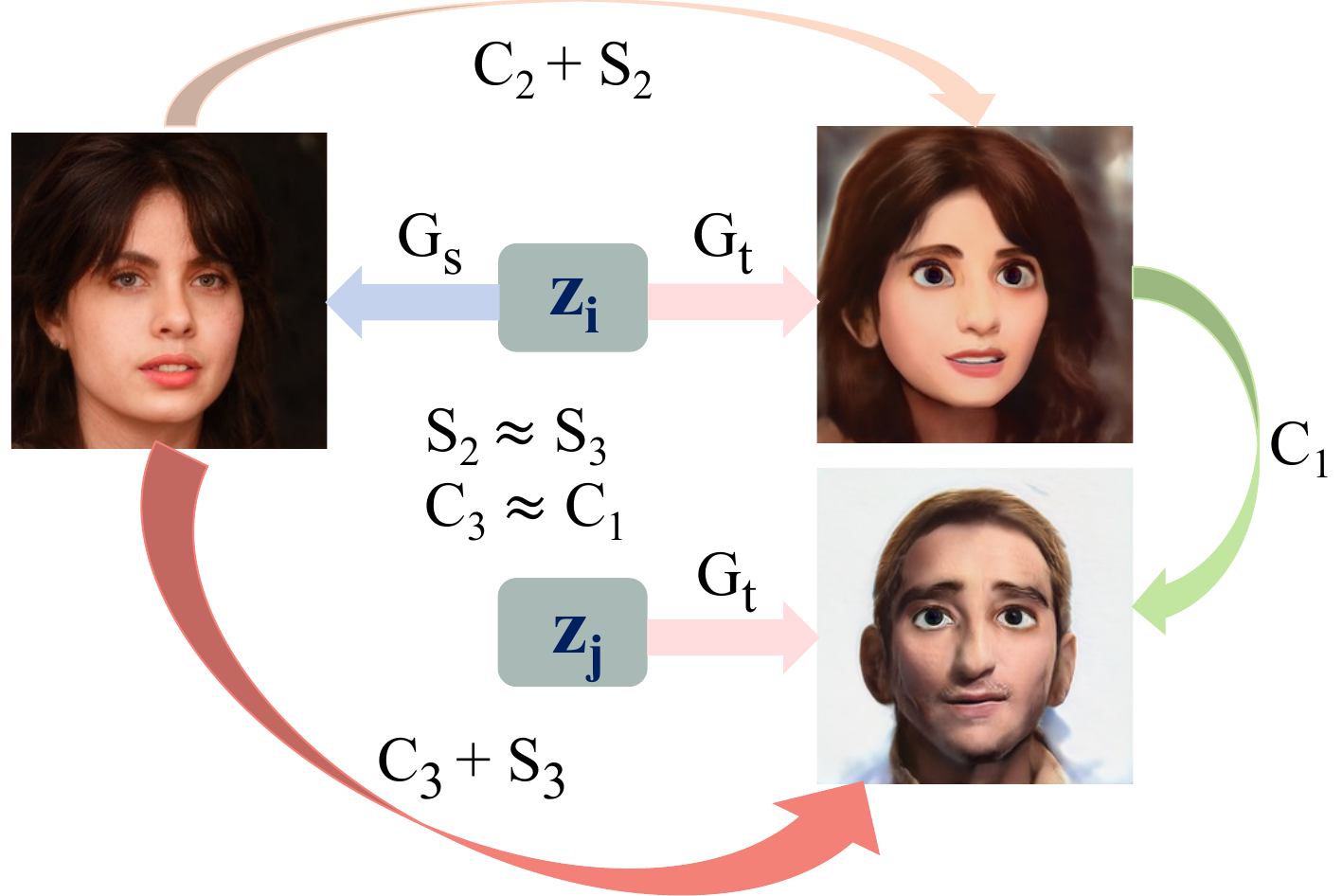}
\end{center}
\vspace{-20pt}
\end{wrapfigure}

We regard the distance $d$  between two images to be the addition of content distance $C$ and style distance $S$, i.e., $d= C+S$, and assume the style distance between the same domain images is 0.
As shown in the right figure, we want $C_1$ to be large, $C_2$ to be small, and minimize the objective function $C_2 - C_1$.
However, directly computing content distance $C_2$ is hard, so we compute $C_2+S_2$ instead. 
Since the style distance between two domains is similar, i.e., $S_2 \approx S_3$,
and content differences between two people should be much larger than that of the same person in different forms, i.e., $C_3 \approx C_1$, the objective function $C_2 - C_1$ becomes $(C_2 + S_2) - (C_1 + S_2) \approx (C_2 + S_2) - (C_3 + S_3)$.
In our problem, the {\it anchor} is $\mathcal{G}_{s}(z_i)$, the {\it positive} example is $\mathcal{G}_{t}(z_i)$, and the {\it negative} example is $\mathcal{G}_{t}(z_j)$.
Then,
our Cross-Domain triplet loss $\mathcal{L}_{cdt}$ is formulated as:
\begin{equation}
  \begin{aligned}
  \mathcal{L}_{cdt} = 
  \mathbb{E}_{\{z_i \sim p_z(z)\}}\max({d^+}(z_i) - {d^-}(z_i) + \alpha, 0)
  \label{eq:cdt_loss}
  \end{aligned}
\end{equation}
\begin{equation}
  \begin{aligned}
  {d^+}(z_i)=\mathcal{L}_{d}(\mathcal{G}_{s}(z_i),\mathcal{G}_{t}(z_i)) 
  \label{eq:cdt_loss1}
  \end{aligned}
\end{equation}
\begin{equation}
  \begin{aligned}
  {d^-}(z_i)=\frac{1}{m-1}\sum_{j, j\neq i}^m \mathcal{L}_{d}(\mathcal{G}_{s}(z_i),\mathcal{G}_{t}(z_j)),
  \label{eq:cdt_loss2}
  \end{aligned}
\end{equation}
where $i$, $j$ are the index of latent codes; ${G}_{s}(z_i)$ is the source model output; ${G}_{t}(z_i)$, ${G}_{t}(z_j)$ are the target model outputs; $m$ is the total number of sampled latent codes; $\alpha$ is the margin; and $\mathcal{L}_{d}$ is a modified LPIPS~\cite{zhang2018unreasonable}
which omits the 4-th layer output of VGG16~\cite{simonyan2014very} in LPIPS.
To validate the design of Cross-Domain triplet loss, we conduct ablation experiments in Sec.~\ref{sec:analysis_cdt}.

{\bf KL-divergence for Adaptive Instance Normalization Inputs.}
The Adaptive Instance Normalization (AdaIN)~\cite{huang2017arbitrary} blocks are important modules in StyleGAN, whose inputs control the ``style'' of the face output. We thus want the AdaIN's inputs of source generator and target generator to share similar distribution.
We propose a KL loss to preserve the relative distance between generations of two latent codes before and after adaptation (similar to~\cite{ojha2021few-shot-gan}), but compute distance on the inputs to AdaIN modules. The KL-AdaIN Loss is formulated as follows:
\begin{equation}
  \mathcal{L}_{kl-adain} = \mathbb{E}_{\{z_i \sim p_z(z)\}}\sum _{l,i}D_{KL}(y^{s,l}_i \parallel y^{t,l}_i)
  \label{eq:kl_adain}
\end{equation}
\begin{equation}
  y^{s,l}_i = {\rm Softmax}(\{{\rm sim}(F^l_s(z_i), F^l_s(z_j))\}_{\forall i \neq j})
  \label{eq:kl_before1}
\end{equation}
\begin{equation}
  y^{t,l}_i = {\rm Softmax}(\{{\rm sim}(F^l_{t}(z_i), F^l_{t}(z_j))\}_{\forall i \neq j}),
  \label{eq:kl_before2}
\end{equation}
where $F_s^l$, $F_t^l$ are the $l$-th AdaIN blocks' inputs; $i, j$ are index of latent codes; $\rm sim$ is the cosine similar function; and $D_{KL}$ indicates KL-divergence.

{\bf Total loss.}
Total loss of our domain adaptation decoder consists of the adversarial loss $\mathcal{L}_{adv}$,  Cross-Domain Triplet loss $\mathcal{L}_{cdt}$ and KL-AdaIN loss $\mathcal{L}_{kl-adain}$:
\begin{equation}
  \begin{aligned}
    \mathcal{L}_{decoder} = & \lambda_{adv} \mathcal{L}_{adv} 
     + \lambda_{cdt}\mathcal{L}_{cdt}
     + \lambda_{kl-adain} \mathcal{L}_{kl-adain},
  \end{aligned}
  \label{eq:decoder_total_loss}
\end{equation}
where $\lambda_*$ are hyper-parameters, $\mathcal{L}_{adv}$ is calculated using an image discriminator, a patch-level discriminator and an ``anchor region'' of latent space to decide which discriminator to use, as introduced in~\cite{ojha2021few-shot-gan}.


\subsection{Style Encoder}
\label{sec:encoder}

To translate a real face photo into an artistic domain, a decent GAN inversion to embed the real face into the latent space is needed.
We aim at {\it learning an encoder that embeds images into the latent space of decoders on different artistic domains}, i.e., the encoder is shared among decoders of different domains.

{\bf $\mathcal{Z+}$ instead of $\mathcal{W}+$.}
Traditional GAN inversion methods mostly embed 
images into the $\mathcal{W}+$ space of StyleGAN~\cite{image2stylegan,richardson2021encoding,e4e}.
The $\mathcal{W}+$ space is designed for better reconstruction. 
After domain adaptation, the encoder’s goal is to find latent codes best suitable for stylization. 
The difference between these two tasks (stylization vs reconstruction) leads to the difference of the most suitable $\mathcal{W}+$:
as shown in Fig.~\ref{fig:compare_encoders}(ii-iii), combining reconstruction-based $\mathcal{W}+$ encoders with our transferred decoder causes some artifacts. 
In our decoder adaptation, the input to the source and target decoders is the same $z$ latent code, so $\mathcal{Z}$ space remains the same after adaptation and is suitable for inversion.
But directly embedding image into $\mathcal{Z}$ space is difficult and sometimes can't maintain some key information, we follow~\cite{agilegan} to extend $\mathcal{Z}$ space to $\mathcal{Z}+$,
by stacking $n$ different $z$ vectors, one for each layer of StyleGAN2 ($n=18$ for $1024^2$, $14$ for $256^2$).

{\bf Architecture.}
The encoder is divided into two parts as in Fig.~\ref{fig:pipeline}: a feature extractor and a sub-encoder.
The feature extractor extracts multi-level features, and the sub-encoder maps multi-level features into $z+$ latent code.
1) Feature Extractor: Since FPN has excellent performance in extracting multi-level features\cite{Lin2017FPN}, and we adopt the FPN as our feature extractor.
2) Sub-encoder: To map the features into latent space, pSp~\cite{richardson2021encoding} uses simple linear layers, we argue that the linear layers have weak performance when there is a big gap between the latent space and features.
Recently, some image segmentation works~\cite{carion2020end,sun2021rethinking} demonstrate that transformer block has excellent ability to translate features (extracted by CNN) to semantic information, we also found that adopting a transformer block instead of linear layers in the sub-decoder helps improve our style encoder's performance (the experiment is in appendix).

{\bf Dual Path Training.}
We utilize a pretrained StyleGAN2 on FFHQ as the decoder and fix its weights during training our encoder.
We observe that during StyleGAN training and domain adaptation, the $z$ latent code is sampled from Gaussian distribution,
so the ideal $\mathcal{Z}+$ space latent codes should also follow Gaussian distribution.
The hVAE in AgileGAN\cite{agilegan} leverages a variational loss to ensure the output latent codes to follow the Gaussian distribution. However, we found it inferior in reconstruction task (Fig.~\ref{fig:compare_encoders}(iv)).
We constrain the encoder output to follow Gaussian distribution by dual path training (Fig.~\ref{fig:pipeline}).
In path-1, a real face photo is fed into our encoder and then the decoder to reconstruct the input face, and we constrain the reconstructed face to be similar to the input face.
In path-2, a $z$ code is sampled, then extended to a $z+$ code by simply repeating $n$ times, and fed into the decoder to output a synthesized face (the process is denoted as path-2a); and the synthesized face then goes through path-1 (denoted as path-2b).
We constrain both the reconstructed $z+$ code and reconstructed image.
Since path-2 contains a cycle, we name it {\it cycle path}.

For path-1, we use $\mathcal{L}_2$ loss, LPIPS~\cite{zhang2018unreasonable} loss $\mathcal{L}_{lpips}$, identity loss $\mathcal{L}_{iden}$ based on ArcFace~\cite{deng2019arcface}, and regularization loss $\mathcal{L}_{reg}$, with the same setting as~\cite{richardson2021encoding}. The loss function is formulated as:
\begin{equation}
  \mathcal{L}_{path1} = \lambda_{L_2}\mathcal{L}_{L_2} + \lambda_{lpips}\mathcal{L}_{lpips} + \lambda_{reg}\mathcal{L}_{reg} + \lambda_{iden}\mathcal{L}_{iden}.
  \label{eq:psploss}
\end{equation}

For path-2, we use the smooth $L_1$ loss~\cite{girshick2015fast} to measure the difference between original $\mathcal{Z}+$ latent codes and embedded ones,  which can be described as:
\begin{equation}
  \mathcal{L}_{z\_predict} = \mathcal{L}_{smooth-L_1}(z_{o}, z_{e}),
  \label{eq:z_predict}
\end{equation}
where $z_{o}$ is the latent code sampled from Gaussian distribution, and $z_{e}$ is the output of our encoder.
We use the same reconstruction loss as path-1 to supervise the image reconstruction in path-2.
The overall loss for training path-2 is:
\begin{equation}
  \mathcal{L}_{path2} = \lambda_{path1}\mathcal{L}_{path1} + \lambda_{{z\_predict}}\mathcal{L}_{z\_predict},
  \label{eq:z_path2}
\end{equation}
where the $\lambda_{*}$  are hyper-parameters.
Dual path is a regularization term, and we conduct a t-SNE~\cite{van2008visualizing} experiment in appendix to show its effects.


\section{Experiment}
\label{sec:experiment}
We implement the proposed method in PyTorch. All experiments are run on a PC with an NVIDIA RTX 3090 GPU.
We use the StyleGAN model pretrained on faces images with $256\times256$ resolution.
We conduct extensive qualitative, quantitative comparison and a perceptual study to demonstrate that the proposed method outperforms state-of-the-arts in artistic portrait generation on various styles under 10-shot and 1-shot settings.
Then, we conduct ablation studies to show the power of the three most important components and analysis of cross-domain triplet loss.
More results and network details are presented in the appendix.

\subsection{Training Details, Datasets and Metrics}
\label{sec:dataset_metrics}

{\bf Training details and Datasets.}
The decoder and encoder are trained separately. 
The encoder is trained only once, and shared among multiple adapted decoders, while one decoder is adapted for each artistic domain.

{\bf Stage-1: encoder training.} To train our style encoder, we use a fixed StyleGAN2 trained on FFHQ dataset~\cite{stylegan} as the decoder, and train the encoder on FFHQ by optimizing loss function $\mathcal{L}_{path1}$ and $\mathcal{L}_{path2}$.
We then use the CelebA-HQ dataset~\cite{karras2017progressive} as test data.

\begin{figure*}[t]
  \centering
  \includegraphics[width=0.95\linewidth]{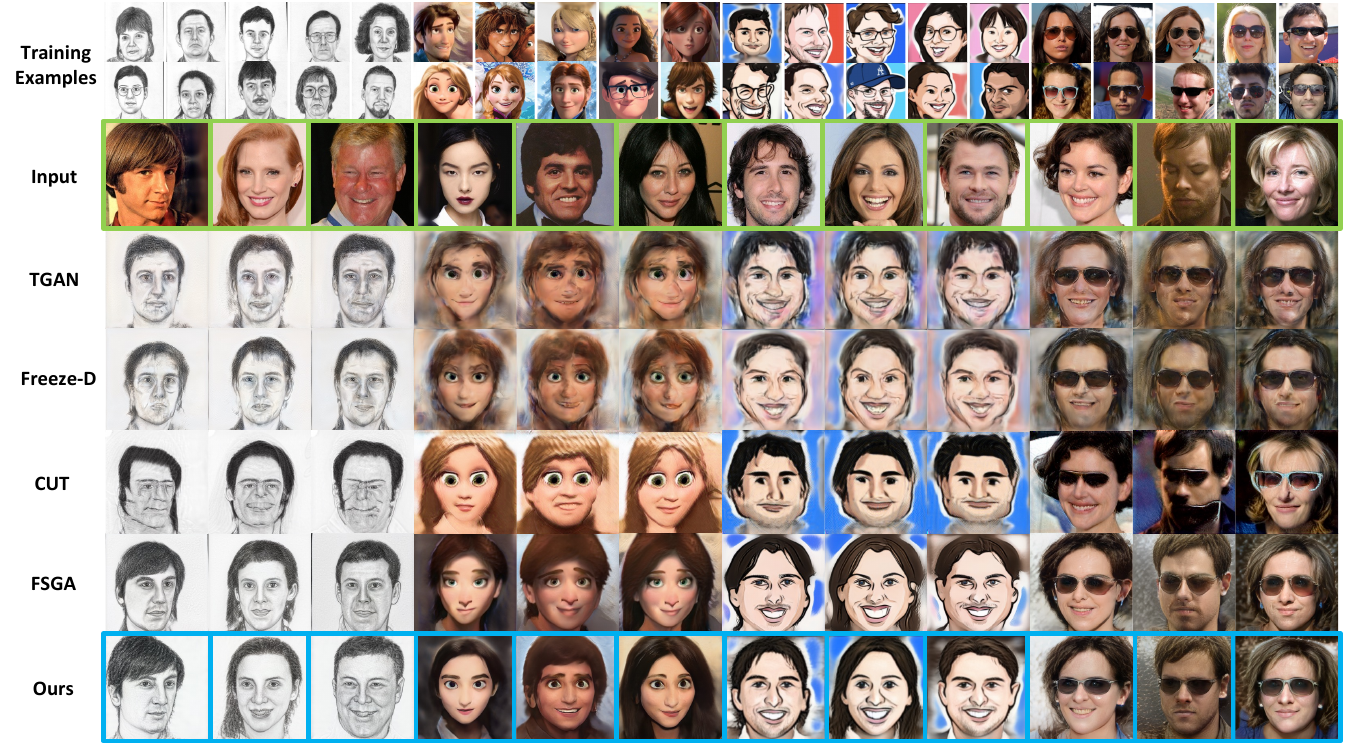}
  \vspace{-0.1in}
  \caption{
  Comparison with FSGA\cite{ojha2021few-shot-gan}, TGAN\cite{wang2018transferring}, Freeze-D\cite{mo2020freeze}, CUT\cite{park2020contrastive} under 10 training images setting.}
  \vspace{-0.15in}
  \label{fig:decoder_compare}
\end{figure*}

{\bf Stage-2: decoder adapting.}
For the few-shot domain adaptation, we use the pretrained StyleGAN2 as the base model, and adapt the base decoder from source domain to a target artistic domain by optimizing $\mathcal{L}_{decoder}$ in Eq.~\ref{eq:decoder_total_loss}.
We adapt one decoder for each of the following target artistic domains, {\it all using no more than 10 artistic portraits images as training data} (Fig.~\ref{fig:decoder_compare}): 1) Sketches from CUHK face sketch dataset\cite{wang2008face}; 2) Caricature from web; 3) Cartoon from Toonify cartoon dataset\cite{pinkney2020resolution}; 4) Raphael from Artistic-Faces\cite{yaniv2019face}; 5) Roy Lichtenstein from Artistic-Faces. We further extend to 6) Sunglasses from FFHQ datset.

{\bf Metrics.}
To quantitatively evaluate our method, we randomly sample 5000 images from CelebA-HQ dataset~\cite{karras2017progressive} as test data and evaluate the generated images using the following three metrics:
{\bf 1) Fréchet Inception Distance (FID)\cite{heusel2017gans}} 
is a widely used metric to evaluate the similarity between the distribution of generated images and the distribution of real data. Lower FID indicates higher similarity and better generation. 
We use FID to measure the distribution similarity between the generated artistic portrait images and the real artistic data. Real data source: for sketch, we use 295 face sketches from CUHK face sketch dataset; for cartoon, we use 252 cartoons from Toonify dataset and web; for sunglasses, we use 2,683 sunglasses images from FFHQ.
{\bf 2) LPIPS Distance~\cite{zhang2018unreasonable}}
is a widely used metric to evaluate the perceptual similarity between two images.
We calculate the LPIPS score between the input face photo and the generated artistic image.
We use this metric to evaluate the identity preservation of generated results.
{\bf 3) LPIPS Cluster:}
Deep models could easily overfit under few-shot setting. Following~\cite{ojha2021few-shot-gan}, we measure the overfitting extent of the generation model using the Intra-cluster pairwise LPIPS distance, which we abbreviate as LPIPS cluster.
The metric assigns the generated images to the nearest training image (by LPIPS distance) and obtains 10 clusters, and calculates the average intra-cluster pairwise distance. 
The higher the average distance, the lower the overfitting degree.

\begin{table}[t]
\begin{center}
\caption{Quantitative comparison with different domain adaptation and unpaired Image-to-Image Translation methods on multiple domains. All methods are trained with 10 images. The metrics are computed from 5,000 generated images.}
\label{tab:quantitative_decoder}
\footnotesize
\scalebox{0.95}{
\begin{tabular}{lccccccccc}
\hline
Models & \multicolumn{3}{c}{Sketches} & \multicolumn{3}{c}{Cartoon}  & \multicolumn{3}{c}{Sunglasses}  \\
& FID$\downarrow$ & Ld$\downarrow$& Lc$\uparrow$ & FID$\downarrow$ & Ld$\downarrow$ & Lc$\uparrow$& FID$\downarrow$ & Ld$\downarrow$ & Lc$\uparrow$ \\
\hline
TGAN~\cite{wang2018transferring} & 134.21 & 0.71 &0.285/0.01 & 199.52 &0.61 & 0.431/0.01 & 88.98 &0.60 & 0.443/0.03\\
Freeze-D~\cite{mo2020freeze} & 132.92 & 0.72 &0.294/0.01& 176.06 & 0.62 &0.437/0.02& 87.94 &0.61 & 0.409/0.02\\
CUT~\cite{park2020contrastive} & 82.21 &0.66 &0.405/0.09 & 176.96 &0.56 & 0.431/0.04 & 85.65 &{\bf 0.45} &{\bf 0.587/0.02}\\
FSGA~\cite{ojha2021few-shot-gan} & 52.35 & 0.69 &0.322/0.02 &104.07 & 0.58 & 0.460/0.03 & 61.40 &0.51 & 0.475/0.02 \\
\hline
Ours & {\bf 49.85} &{\bf 0.58} & {\bf 0.424/0.03} & {\bf 84.93} & {\bf 0.51} & {\bf 0.515/0.03} &{\bf 48.29} &0.50 &0.482/0.01 \\
\hline
\vspace{-0.15in}
\end{tabular}
}
\end{center}
\end{table}

\subsection{Comparisons with Few-Shot Generation Models}
\label{sec:comparison}

{\bf Comparison Methods.} We compare our few-shot domain adaptation decoder with domain adaptation methods and an unpaired Image-to-Image Translation method {\it under few-shot setting (10 training images)}: 
(i) Few-Shot-GAN-Adaptation (FSGA)~\cite{ojha2021few-shot-gan}: adapts a pretrained model in source domain to target domain via cross-domain correspondence. 
(ii) TGAN~\cite{wang2018transferring}: transfers a pretrained source domain model to target domain by finetuning the original loss function. 
(iii) Freeze-D~\cite{mo2020freeze}: freezes the first three layers of the discriminator during adaptation. 
(iv) CUT~\cite{park2020contrastive}: an unpaired Image-to-Image Translation method based on patch-wise contrastive learning.
We use the author implementations for (i), (iii), (iv) and implement (ii) by ourselves.
For (ii)(iii), we use the patch-wise discriminator the same as (i) and Ours to ensure a fair comparison.

For encoder, we compare our encoder with (i) pSp~\cite{richardson2021encoding} and e4e~\cite{e4e} encoder, which encode real face photos into $\mathcal{W}+$ space; 
and (ii) the hVAE $\mathcal{Z}+$ encoder in AgileGAN~\cite{agilegan}.
We use author implementations for (i) and since (ii) AgileGAN is not open-sourced, we implement its encoder following the paper description.

{\bf Qualitative Comparison.}
Fig.~\ref{fig:decoder_compare} shows qualitative comparisons with different domain adaptation methods and unpaired Image-to-Image Translation methods on multiple target domains, i.e., Sketches, Cartoon, Caricature, and Sunglasses.
Results of CUT show clear overfitting, except sunglasses domain; FreezeD and TGAN results contain cluttered lines in all domains; Few-Shot-GAN-Adaptation results preserve the identity but still show overfitting; while our results well preserve the input facial features, show the least overfitting, and significantly outperform the comparison methods on all four domains.

We show qualitative comparisons with different encoders in Fig.~\ref{fig:compare_encoders}.
Results of pSp and e4e show clear artifacts in generation results, and hVAE $\mathcal{Z}+$ encoder is worse in identity preservation.
In contrast, our encoder has good performance in reducing artifacts and outperforms the comparison encoders.

\begin{table}[t]
	\begin{minipage}[t]{0.5\linewidth}
        \centering
        \footnotesize
        \caption{Quantitative comparison with different encoders.}
        \label{tab:quantitative_encoder}
        \scalebox{0.75}{
        \begin{tabular}{lcccccc}
        \hline
        Encoders &  \multicolumn{2}{c}{Sketches} & \multicolumn{2}{c}{Cartoon}  & \multicolumn{2}{c}{Sunglasses} \\
        &FID $\downarrow$& Ld$\downarrow$ &FID $\downarrow$& Ld$\downarrow$&FID$\downarrow$ & Ld$\downarrow$\\
        \hline
        pSp~\cite{richardson2021encoding} & 90.96 &0.597 & 92.22 &0.565 & 66.08& 0.531\\
        e4e~\cite{e4e} & 101.19 & 0.596 &93.37 & 0.542 &62.03& 0.528\\
        hVAE $\mathcal{Z}+$~\cite{agilegan} & 51.49 & 0.589 &  94.63 & 0.554 & 57.39& {\bf 0.504}  \\
        \hline
        Ours  &  {\bf 49.85}& {\bf 0.586} & {\bf 90.80 }& {\bf 0.537} & {\bf 48.29} & {\bf 0.504}\\    
        \hline
        \end{tabular}
        }
        \vspace{-0.1in}
	\end{minipage}
	\begin{minipage}[t]{0.5\linewidth}
        \centering
        \footnotesize
        \caption{User study results.}
        \label{tab:userstudy}
        \scalebox{0.92}{
        \begin{tabular}{lccccc}
        \hline
        & Ours & FSGA & Freezed & TGAN & CUT  \\
        \hline
        Rank1 &{\bf 59.6$\%$} & 25.7$\%$ & 2.9$\%$  & 3.7$\%$  & 8.2$\%$ \\
        Rank2 &26.2$\%$ & 56.4$\%$ & 5.3$\%$  & 5.3$\%$  & 6.8$\%$ \\
        Rank3 &7.6$\%$ & 9.7$\%$ & 36.6$\%$ & 26.1$\%$ & 20.0$\%$ \\
        Rank4 &3.5$\%$  & 4.3$\%$&  34.2$\%$ & 45.7$\%$ & 12.3$\%$ \\
        Rank5 &3.1$\%$  & 3.9$\%$&  21.0$\%$ & 19.3$\%$ & 52.7$\%$ \\
        \hline
        \end{tabular}
        }
	\end{minipage}
\end{table}

\begin{figure}[t]
	\centering
    \includegraphics[width=\linewidth]{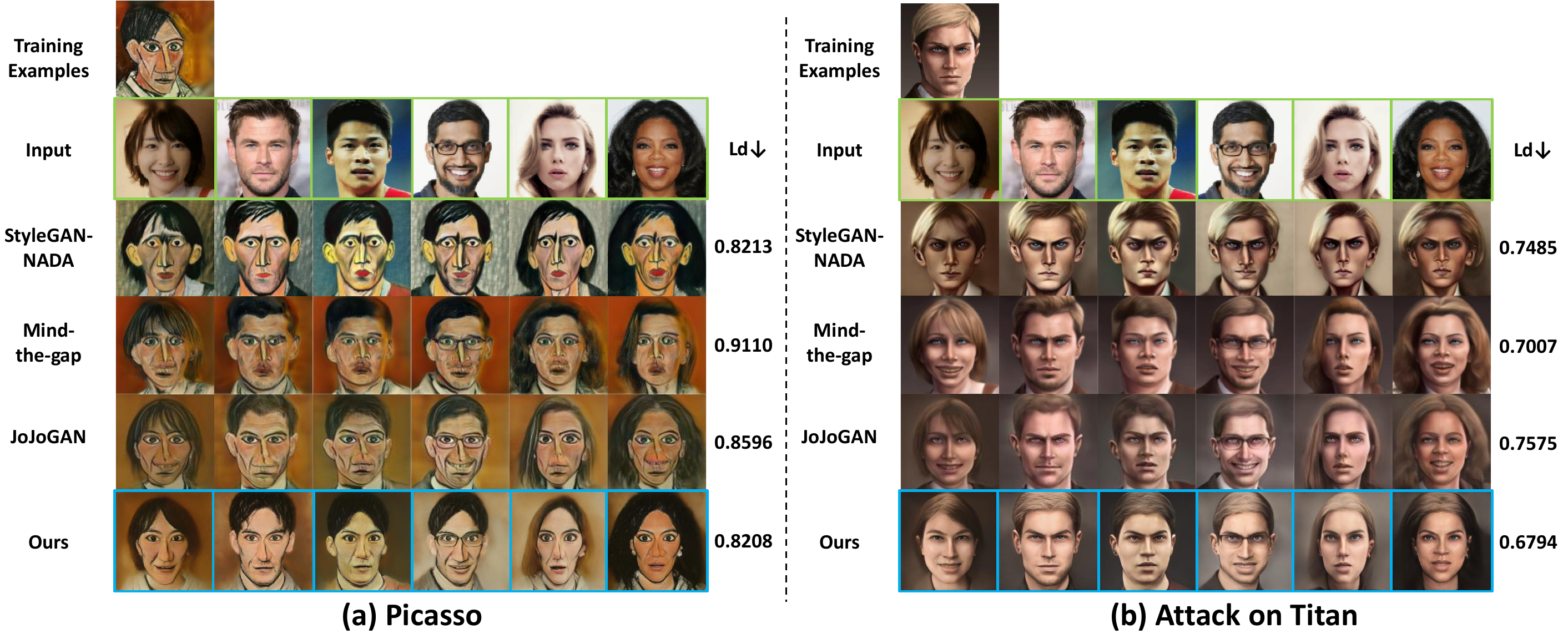}
    \vspace{-0.25in}
    \caption{
	Comparison with StyleGAN-NADA\cite{abdal2021styleflow}, Mind-the-gap\cite{zhu2021mind}, JoJoGAN\cite{chong2021jojogan} under one-shot setting. All models utilize a pretrained StyleGAN, and StyleGAN-NADA, Mind-the-gap leverage external knowledge in a pretrained CLIP. Ld denotes LPIPS distance to input photos.
	}
	\label{fig:one-shot}
\end{figure}

\begin{figure*}
  \centering
  \includegraphics[width=1\linewidth]{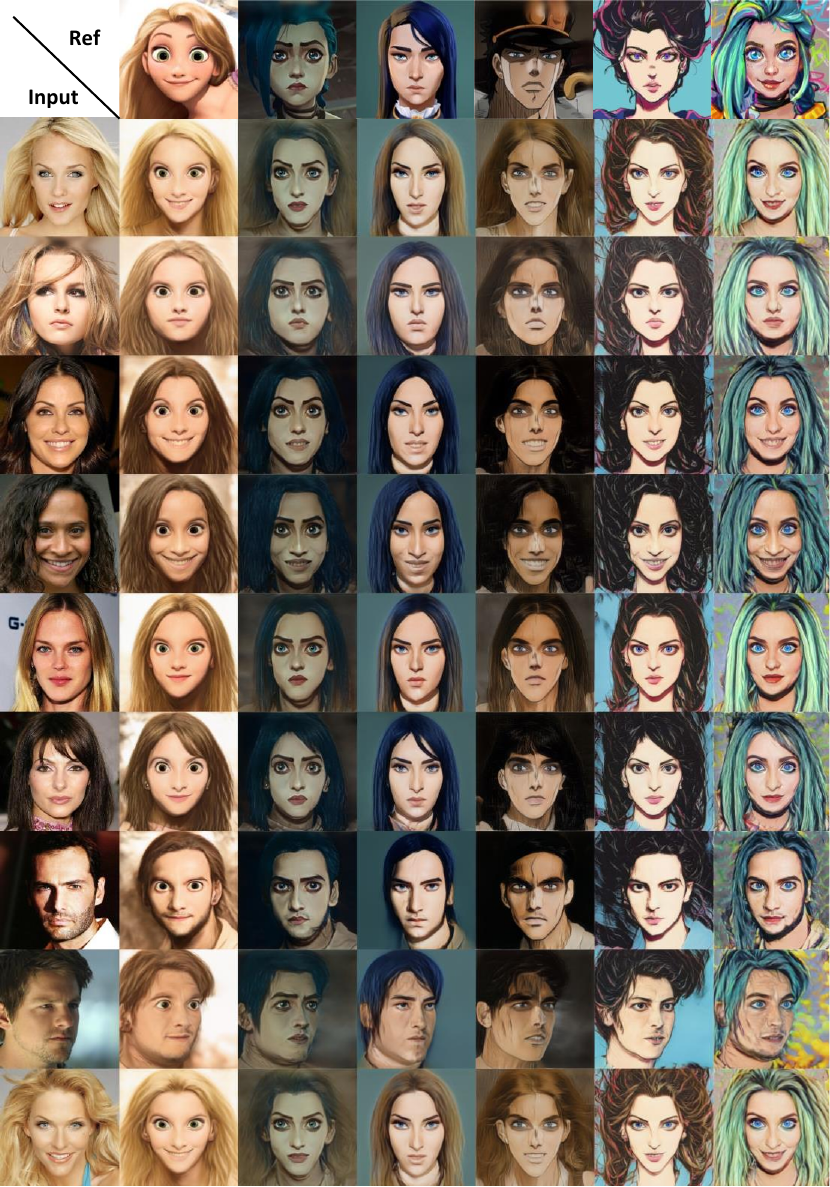}
  \vspace{-0.1in}
  \caption{\captionOneShot}
  \vspace{-0.1in}
  \label{fig:more_oneshot1}
\end{figure*}
\begin{figure*}
  \centering
  \includegraphics[width=1\linewidth]{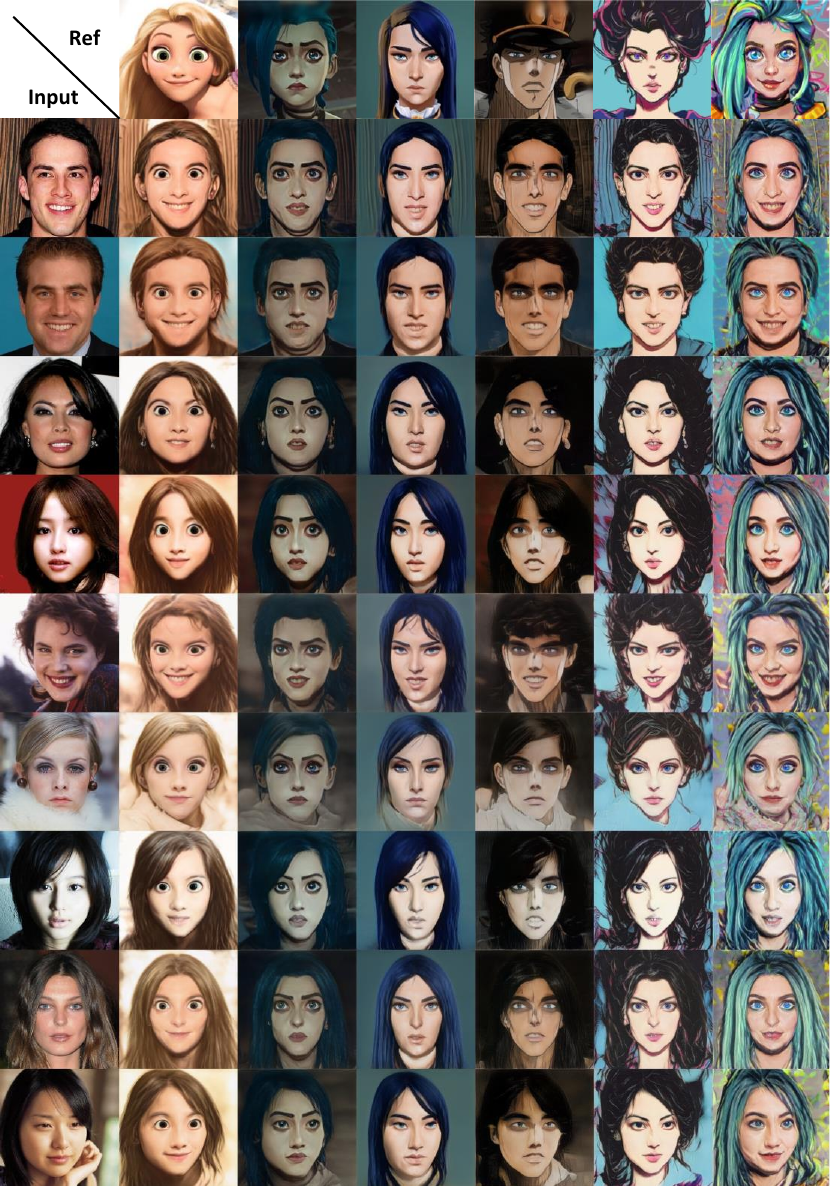}
  \vspace{-0.1in}
  \caption{\captionOneShot}
  \vspace{-0.1in}
  \label{fig:more_oneshot2}
\end{figure*}
\begin{figure*}
  \centering
  \includegraphics[width=1\linewidth]{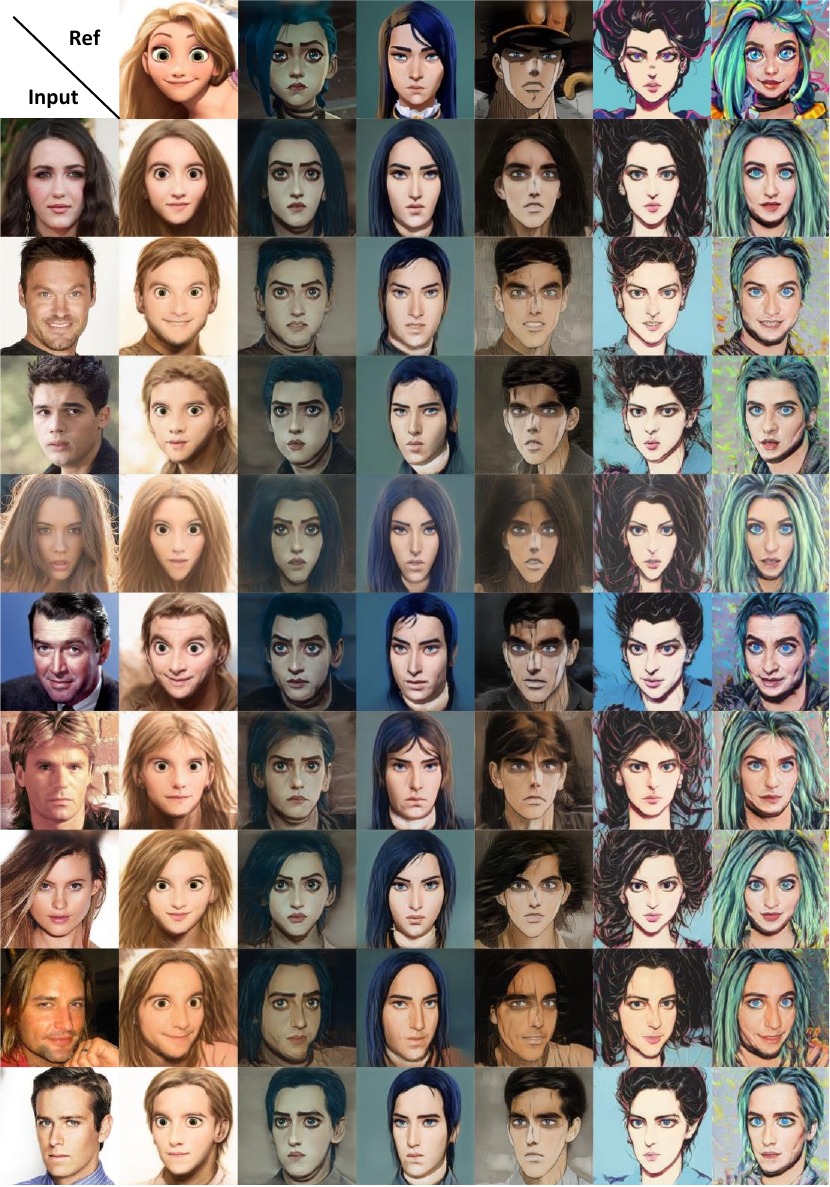}
  \vspace{-0.1in}
  \caption{\captionOneShot}
  \vspace{-0.1in}
  \label{fig:more_oneshot3}
\end{figure*}

{\bf Quantitative Comparison.}
Table~\ref{tab:quantitative_decoder} shows the FID, LPIPS distance (Ld), and LPIPS cluster (Lc) scores of ours and different domain adaptation methods and unpaired Image-to-Image Translation methods on multiple target domains, i.e., Sketches, Cartoon and Sunglasses.
Our few-shot domain adaptation decoder achieves the best FID on all three domains. We also achieve the best LPIPS distance and LPIPS cluster on Sketches and Cartoon domain. 
For Sunglasses domain, our LPIPS distance and LPIPS cluster are worse than CUT, but qualitative results (Fig.~\ref{fig:decoder_compare}) show CUT simply blackens the eye regions and leads to obvious artifacts.

\begin{figure}[t]
\subfigure[Ablation study]{
	\begin{minipage}[t]{0.48\linewidth}
        \centering
        \includegraphics[width=1.0\linewidth]{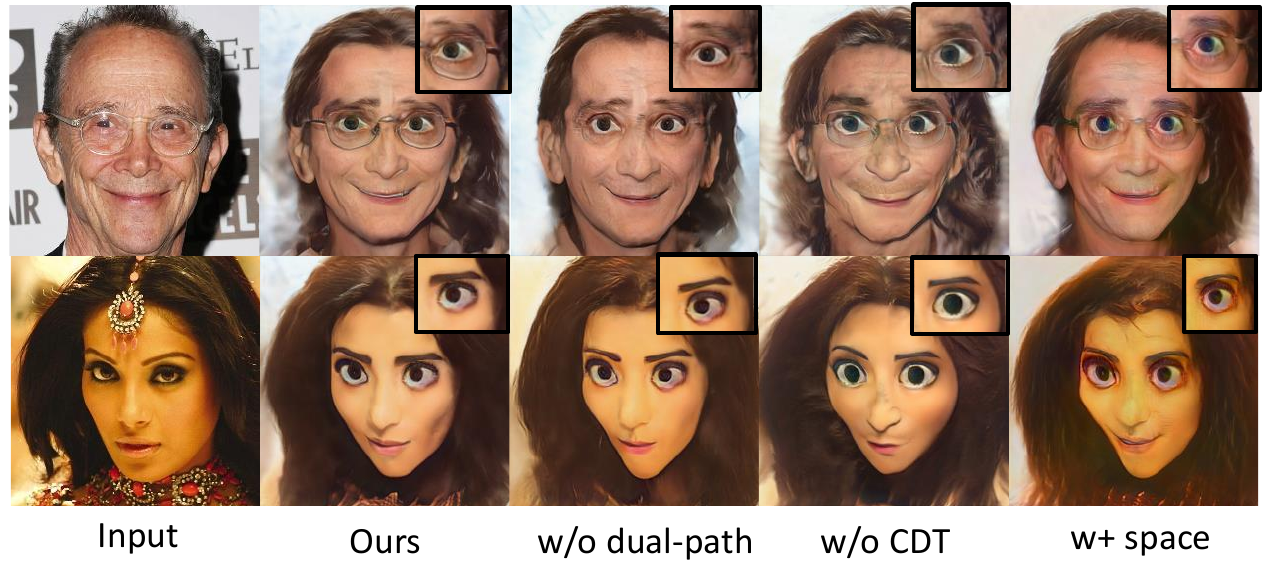}
        \vspace{-0.2in}
        \vspace{-0.2in}
        \label{fig:ablation_main}
	\end{minipage}
	}
\subfigure[Analysis of CDT]{
	\begin{minipage}[t]{0.48\linewidth}
        \centering
        \includegraphics[width=1.0\linewidth]{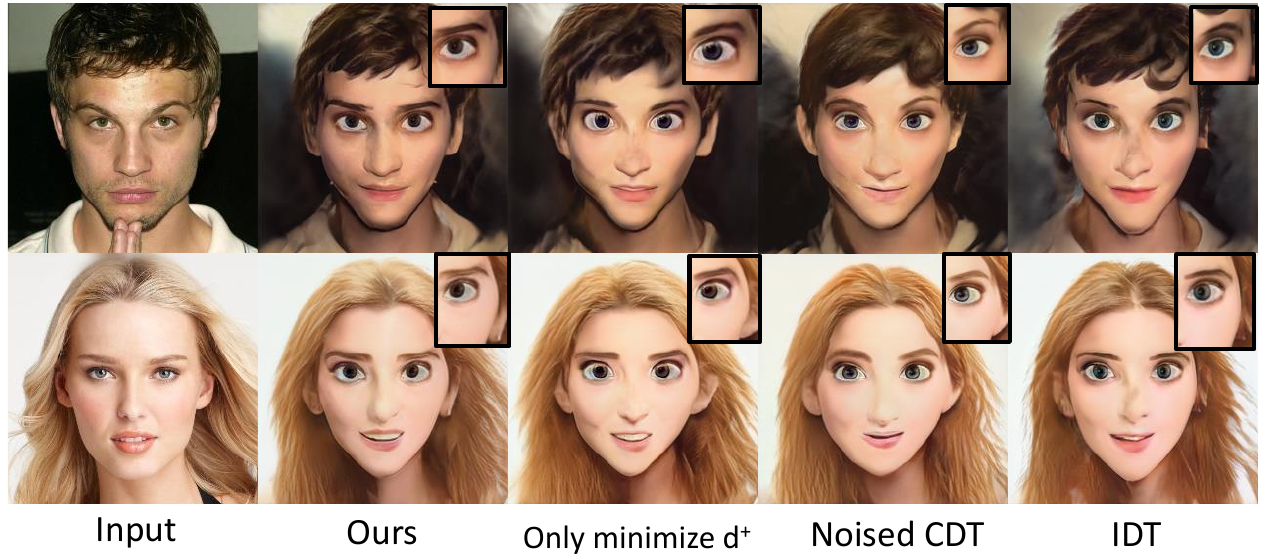}
        \vspace{-0.2in}
        \vspace{-0.2in}
        \label{fig:analysis_cdt}
	\end{minipage}
	}
	\vspace{-0.2in}
	\caption{(a) Ablation study on three key components;(b)Analysis of Cross-Domain Triplet loss.}
\end{figure}

Table~\ref{tab:quantitative_encoder} shows the FID, LPIPS distance of ours and different encoders on multiple target domains, i.e., Sketches, Cartoon and Sunglasses.
Our encoder is better than all three comparison encoders on all three domains. Results show that our encoder better copes with our artistic domain decoder and generates better results in distribution similarity and identity preservation.

{\bf User study.}
We conduct a user study to compare our method with TGAN, Freeze-D, CUT and Few-Shot-GAN-Adaptation in 10-shot setting. 
We randomly sample 120 images from CelebA-HQ dataset, and generate artistic portraits in 4 domains (Sketches, Cartoon, Caricature, Sunglasses).
Participants are required to rank the results of comparison methods and ours considering 
generation quality, style consistency and identity preservation.
62 participants attended the user study and each of them compared 30 groups of results (randomly sampled from 120).
Results of the percentages of each method ranked as 1-5 are summarized in Table~\ref{tab:userstudy}. Our method ranks the best in 59.6\% of votes, which significantly outperforms other methods.
The average rankings of different methods are: ours 1.64, Few-Shot-GAN-Adaptation 2.04, FreezeD 3.65, TGAN 3.72 and CUT 3.94.

\subsection{Comparisons with One-Shot Domain Adaptation Models}
\label{sec:oneshot-comparison}
{\bf Comparison Methods.} Recently, some notable one-shot domain adaptation methods are developed based on pretrained StyleGAN and CLIP models. 
1) StyleGAN-NADA\cite{gal2021stylegan}: leverages the semantic power of CLIP\cite{clip} model and uses text to guide the domain adaptation, its code also implements one-shot domain adaptation based on a reference image; 
2) Mind-the-GAP~\cite{zhu2021mind}: a one-shot domain adaptation method that leverages CLIP model and StyleGAN inversion and computes the domain gap as a direction in CLIP embedding space;
3) JoJoGAN~\cite{chong2021jojogan}: a one-shot domain adaptation method that generates a large dataset from a single reference by mixing style latent codes and then finetunes StyleGAN using pixel-level loss.

In Fig.~\ref{fig:one-shot}, we compare with these methods under one-shot setting on two artistic domains.
StyleGAN-NADA results are worse in style similarity;
Mind-the-gap and JoJoGAN are unstable for some domain (Fig.~\ref{fig:one-shot}(a)), because they first invert the reference image of target domain back to FFHQ faces domain, and this is difficult for abstract style like Picasso.
\ourGAN{} achieves good stylization and has the lowest LPIPS distance (Ld) to input photos.

{\bf More 1-shot results are shown in Figs~\ref{fig:more_oneshot1},~\ref{fig:more_oneshot2},~\ref{fig:more_oneshot3}, including 27 test photos and six different artistic domains, where the training examples are shown in the top row.}

\begin{table}[t]
    \vspace{-0.15in}
	\begin{minipage}[t]{0.48\linewidth}
        \centering
        \scriptsize
        \caption{Ablation study quantitative scores.}
        \label{tab:ablation}
        \begin{tabular}{lccc}
        \hline
        Models & FID $\downarrow$ & Ld $\downarrow$ & Lc $\uparrow$ \\
        \hline
        w/o CDT & 113.56 & 0.56 & 0.498/0.04 \\
        w/o dual-path & 88.89 & 0.51 & 0.498/0.05\\
        $ z+ \to w+ $ & 92.23 & 0.56 & 0.474/0.03\\
        \hline
        Ours & {\bf 84.93} & {\bf 0.51} &{\bf 0.515/0.03}\\
        \hline
        \end{tabular}
	\end{minipage}
	\begin{minipage}[t]{0.48\linewidth}
        \centering
        \scriptsize
        \caption{Analysis of Cross-domain Triplet Loss.}
        \label{tab:decoder_analysis}
        \begin{tabular}{lccc}
        \hline
        Models & FID $\downarrow$ & Ld $\downarrow$ & Lc $\uparrow$ \\
        \hline
        only minimizing $d^+$ & 108.07 & 0.54 & 0.511/0.02 \\
        close $z$ instead of same $z$ & 99.64 & 0.53 & 0.495/0.04\\
        only $C$ instead of $C+S$ & 106.49 & 0.59 & 0.489/0.03\\
        \hline
        Ours  & {\bf 84.93} & {\bf 0.51} & {\bf 0.515/0.03}\\    
        \hline
        \end{tabular}
	\end{minipage}
\end{table}

\subsection{Ablation Study on Three Key Components}
\label{sec:ablation}

We conduct ablation studies on three key components of our method: the cross-domain triplet loss (CDT) in our decoder, the $Z+$ space and the dual training path in our encoder.
We train the ablated models by removing each component and evaluate the metrics. 
As shown in Fig.~\ref{fig:ablation_main} and Table.~\ref{tab:ablation}, each component plays an important role in our final results.

\subsection{Analysis of Cross-Domain Triplet Loss}
\label{sec:analysis_cdt}

To validate the design of cross-domain triplet loss, we conduct comparison with three different designs: (1) only minimizing $d^+$, i.e., the distances between images generated by the same latent code for source and target domain;
(2) using a close $z$ instead of the same $z$ in positive pair (Noised CDT);
(3) only concerning content distance without style distance (In-Domain Triplet Loss, IDT).

As shown in Table.~\ref{tab:decoder_analysis}, our Cross-Domain Triplet loss has better FID, Ld and Lc score than other settings.
As shown in Fig.~\ref{fig:analysis_cdt}, the model trained with our CDT has the best visual quality.

\section{Conclusion and Discussion}
\label{sec:Conclution}
In this paper, we propose CtlGAN, a new framework for few-shot artistic portraits generation (no more than 10 artistic faces). 
With a new contrastive transfer learning strategy, we effectively avoid overfitting in few-shot generation.
And with a new encoder with $Z+$ latent space setting and dual path training, we generate high-quality artistic portraits 
while keeping the identity.
Extensive qualitative, quantitative comparisons and a user study show our method achieves state-of-the-art performance.
Our model mainly targets artistic portraits generation and has some limitations for local editing, as shown in Fig.~\ref{fig:decoder_compare}, the FFHQ$\to$Sunglasses model sometimes changes the haircut and skin details.
In the future, we wish to develop a model suitable for both global style change and local editing.

\begin{subappendices}

\renewcommand{\thesection}{\Alph{section}}
\section{Overview}
This appendix includes:
\begin{itemize}
\item more 1-shot and 10-shot results on multiple artistic domains (Sec.~\ref{sec:more_result});
\item results on other domains (Sec.~\ref{sec:moredomain});
\item more analysis on each component: 1) t-SNE visualization of the dual-path impact (Sec.~\ref{sec:dual-path}); 2) analysis of sub-encoder (Sec.~\ref{sec:analysis_sub-encoder}); 3) more ablation studies on decoder (Sec.~\ref{sec:more_ablation}); 4) detailed analysis on triplet loss (Sec.~\ref{sec:triplet});
\item detailed network architecture, training settings and hyper-parameters (Sec.~\ref{sec:network_detail});
\item comparison with neural style transfer and image-to-image translation methods (Sec.~\ref{sec:campare}).
\end{itemize}

\begin{figure*}[t]
  \centering
  \includegraphics[width=\linewidth]{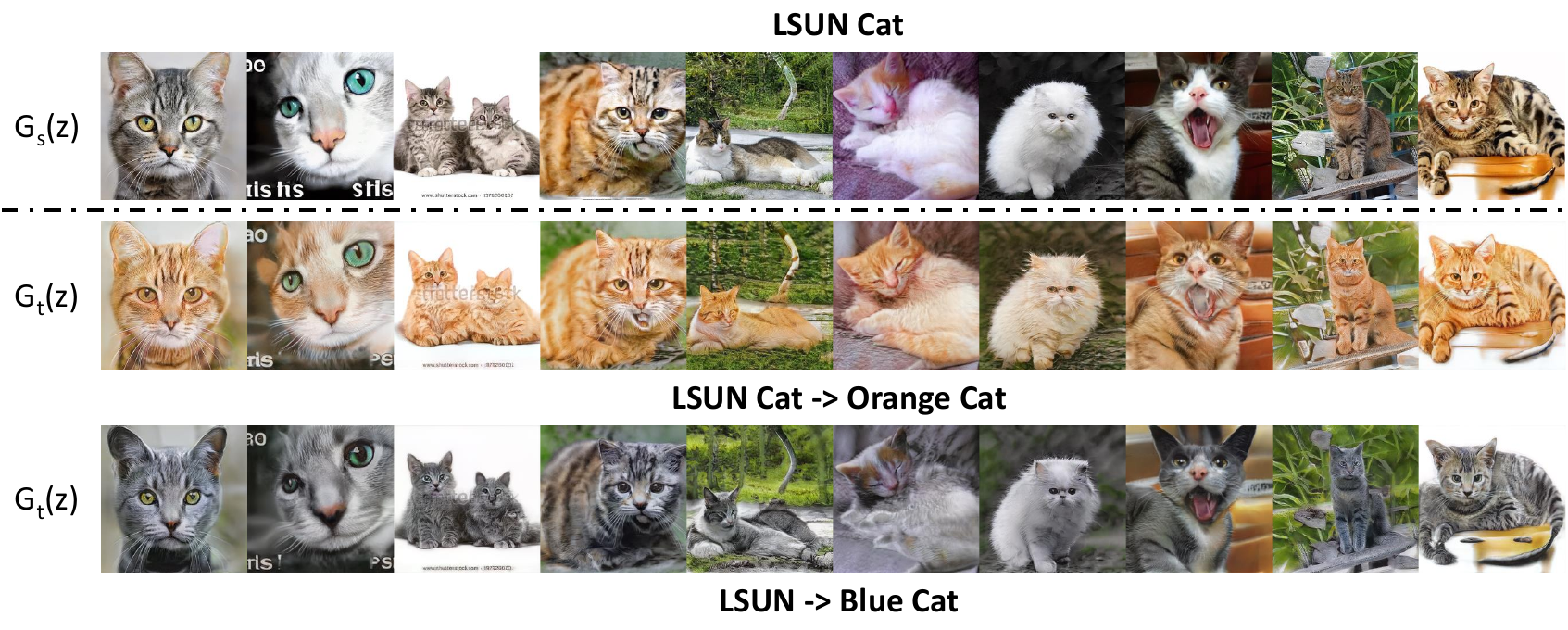}
  \vspace{-0.1in}
  \caption{Results on Cats. The source domain is LSUN\cite{yu15lsun} Cat, and the target domain is Orange Cat or Blue Cat.}
  \vspace{-0.1in}
  \label{fig:cat}
\end{figure*}

\begin{figure*}[t]
  \centering
  \includegraphics[width=\linewidth]{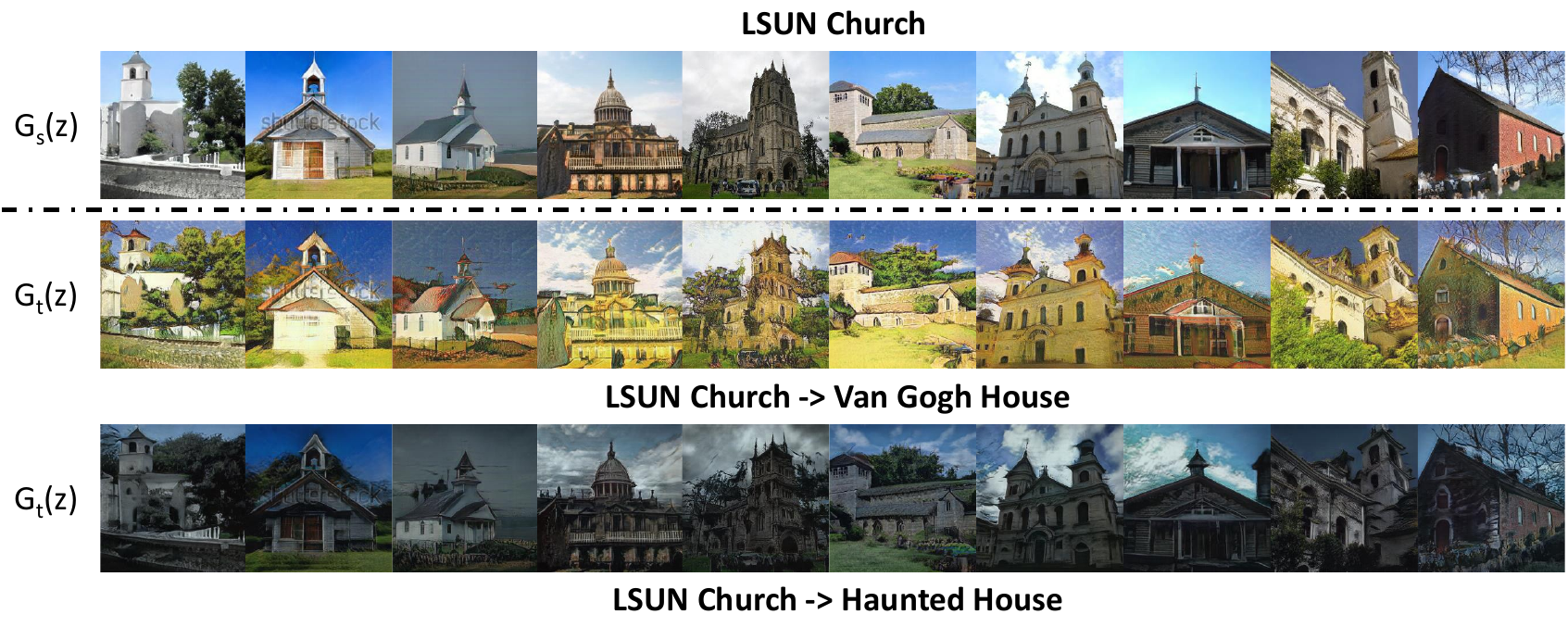}
  \vspace{-0.1in}
  \caption{Results on Churches. The source domain is LSUN\cite{yu15lsun} Church, and the target domain is Van Gogh House or Haunted House.}
  \vspace{-0.1in}
  
  \label{fig:church}
\end{figure*}


\section{1-shot and 10-shot results on multiple artistic domains}
\label{sec:more_result}
In this section, we show more results on multiple artistic domains under 1-shot and 10-shot training.

{\bf 1-shot results} are shown in Figs.~\ref{fig:more_oneshot1},~\ref{fig:more_oneshot2},~\ref{fig:more_oneshot3}, including 27 test photos and six different artistic domains, where the training examples are shown in the top row.

\begin{figure*}
  \centering
  \includegraphics[width=0.95\linewidth]{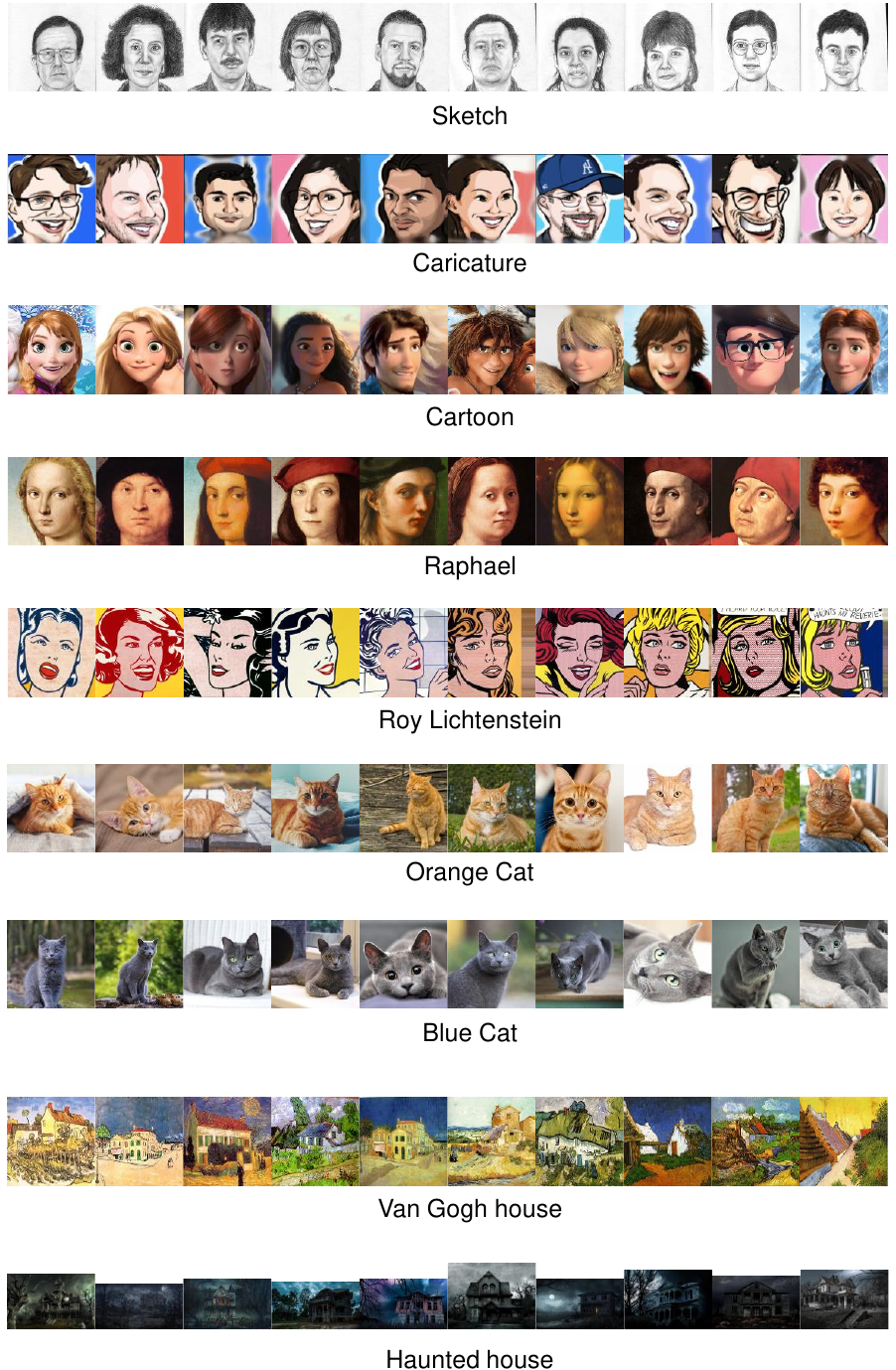}
  \vspace{-0.1in}
  \caption{The 10 training images used for different artistic domains.}
  \vspace{-0.1in}
  \label{fig:training}
\end{figure*}

\begin{figure*}
  \centering
  \small
  \makebox[\MRscale\textwidth]{Inputs}
  \makebox[\MRscale\textwidth]{Sketches}
  \makebox[\MRscale\textwidth]{Caricature}
  \makebox[\MRscale\textwidth]{Cartoon}
  \makebox[\MRscale\textwidth]{Raphael}
  \makebox[\MRscale\textwidth]{RoyLichtenstein}\\
  \normalsize
  \includegraphics[width=0.9\linewidth]{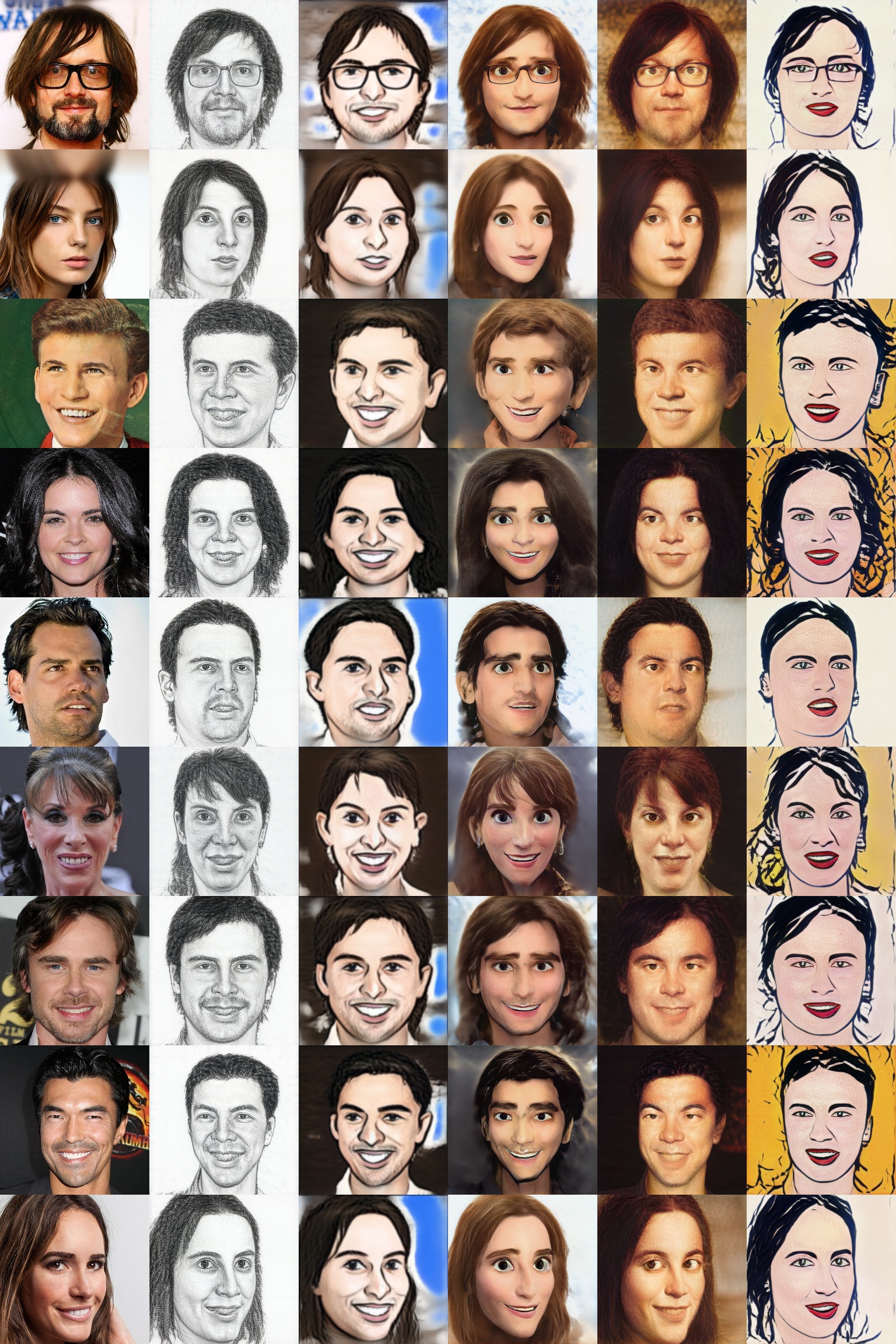}
  \caption{\captionMR{}}
  \label{fig:more_result0}
\end{figure*}

\begin{figure*}
  \centering
  \small
  \makebox[\MRscale\textwidth]{Inputs}
  \makebox[\MRscale\textwidth]{Sketches}
  \makebox[\MRscale\textwidth]{Caricature}
  \makebox[\MRscale\textwidth]{Cartoon}
  \makebox[\MRscale\textwidth]{Raphael}
  \makebox[\MRscale\textwidth]{RoyLichtenstein}\\
  \normalsize
  \includegraphics[width=0.9\linewidth]{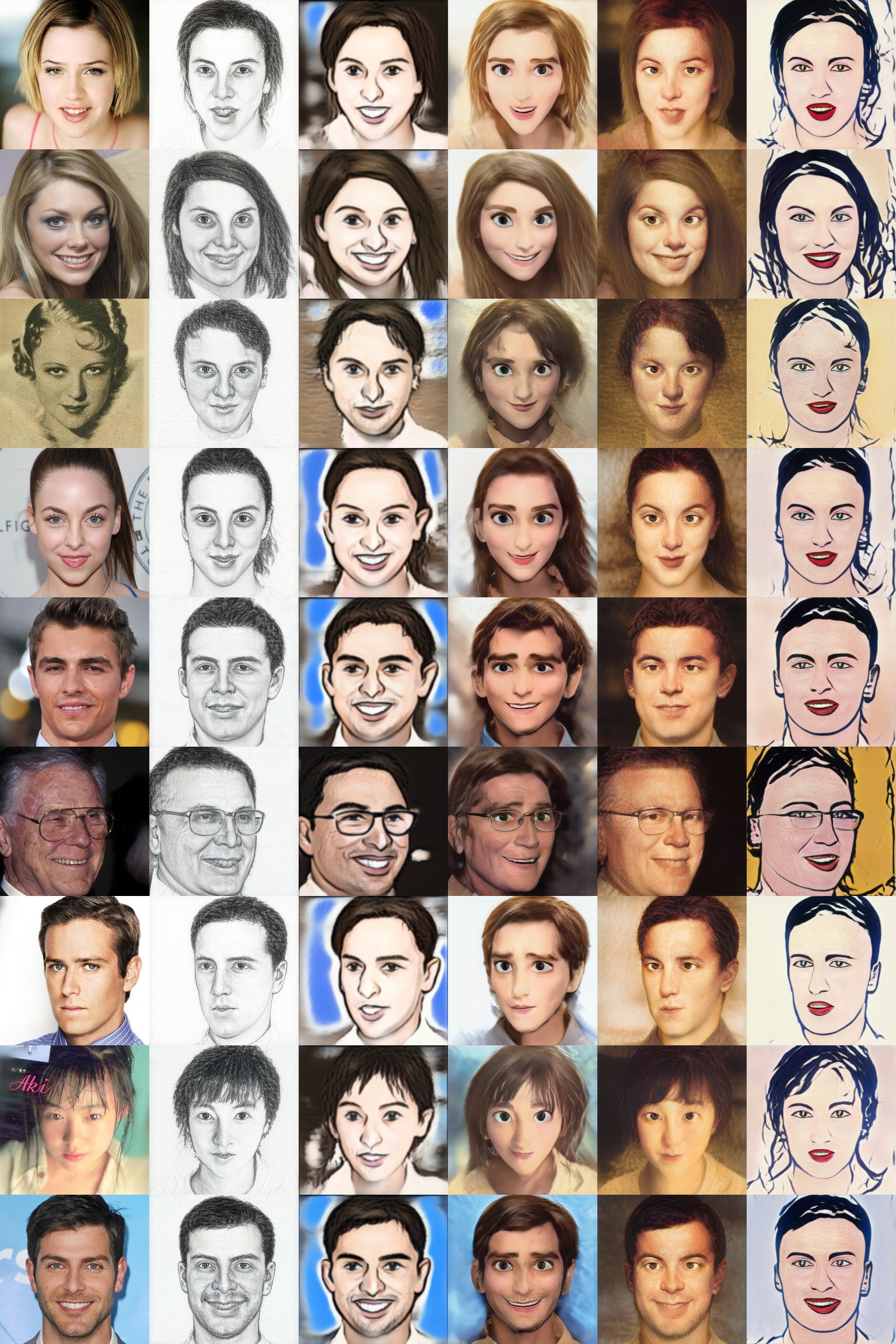}
  \caption{\captionMR{}}
  \label{fig:more_result1}
\end{figure*}

\begin{figure*}
  \centering
  \small
  \makebox[\MRscale\textwidth]{Inputs}
  \makebox[\MRscale\textwidth]{Sketches}
  \makebox[\MRscale\textwidth]{Caricature}
  \makebox[\MRscale\textwidth]{Cartoon}
  \makebox[\MRscale\textwidth]{Raphael}
  \makebox[\MRscale\textwidth]{RoyLichtenstein}\\
  \normalsize
  \includegraphics[width=0.9\linewidth]{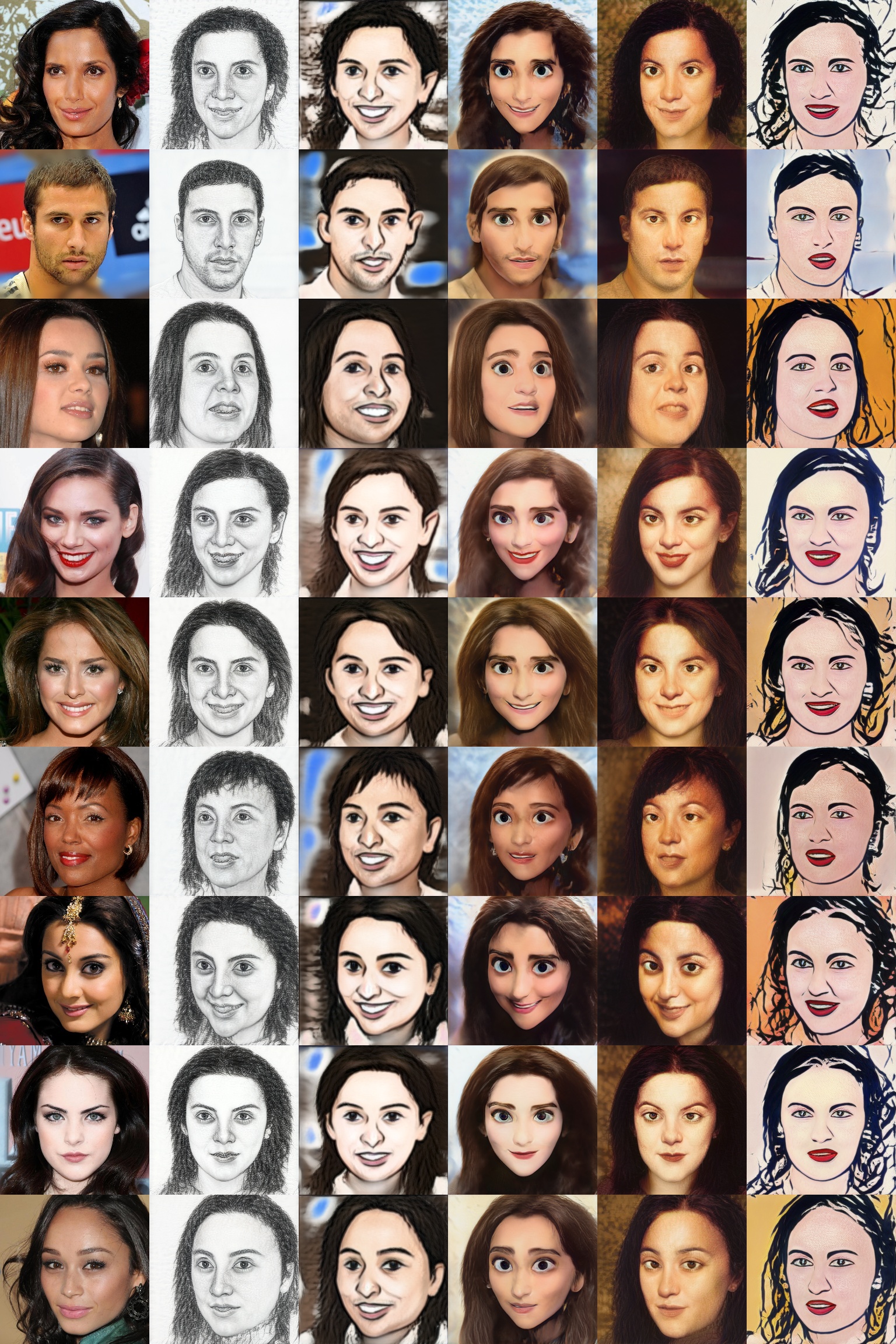}
  \caption{\captionMR{}}
  \label{fig:more_result2}
\end{figure*}

\begin{figure*}
  \centering
  \small
  \makebox[\MRscale\textwidth]{Inputs}
  \makebox[\MRscale\textwidth]{Sketches}
  \makebox[\MRscale\textwidth]{Caricature}
  \makebox[\MRscale\textwidth]{Cartoon}
  \makebox[\MRscale\textwidth]{Raphael}
  \makebox[\MRscale\textwidth]{RoyLichtenstein}\\
  \normalsize
  \includegraphics[width=0.9\linewidth]{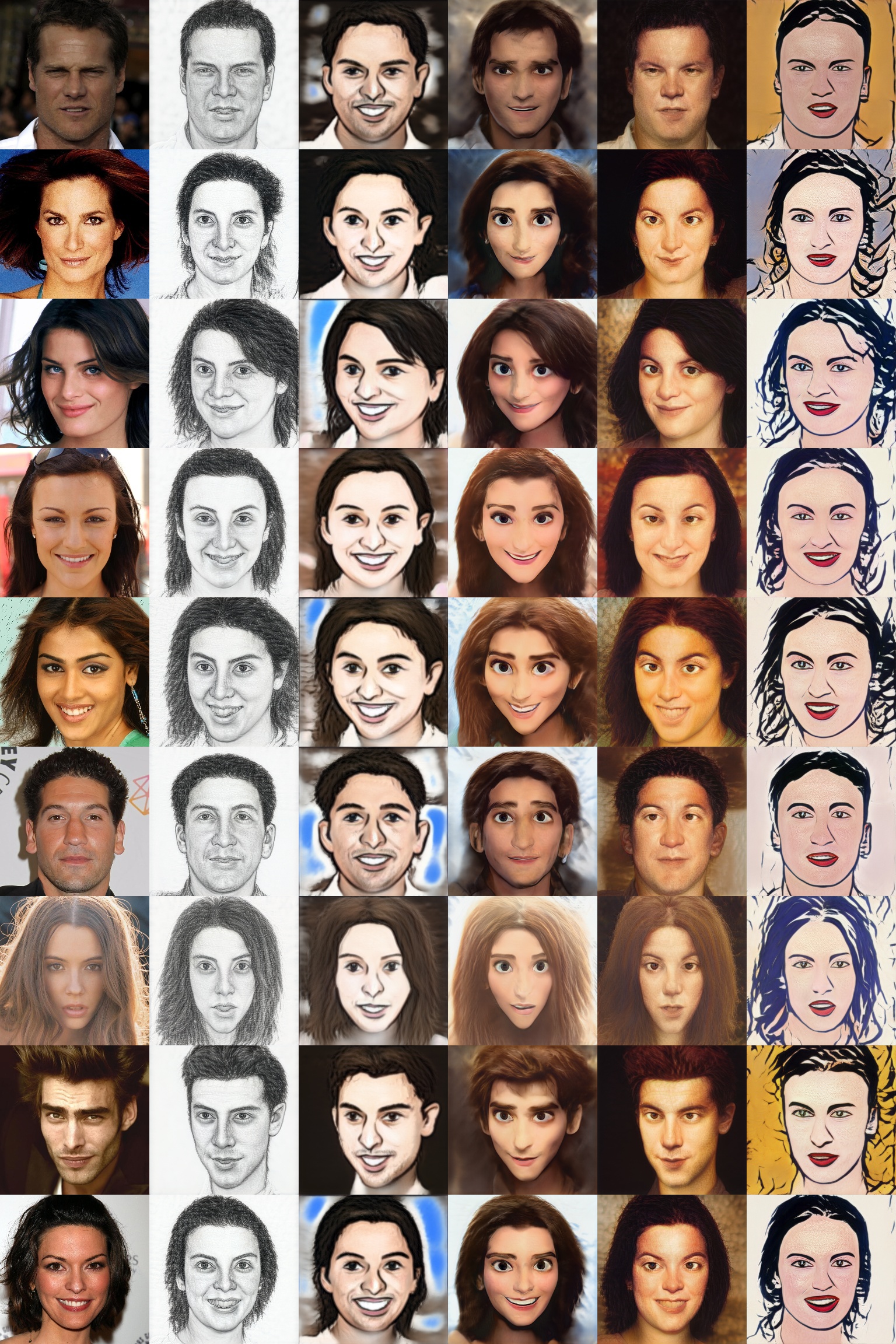}
  \caption{\captionMR{}}
  \label{fig:more_result3}
\end{figure*}

\begin{figure*}
  \centering
  \small
  \makebox[\MRscale\textwidth]{Inputs}
  \makebox[\MRscale\textwidth]{Sketches}
  \makebox[\MRscale\textwidth]{Caricature}
  \makebox[\MRscale\textwidth]{Cartoon}
  \makebox[\MRscale\textwidth]{Raphael}
  \makebox[\MRscale\textwidth]{RoyLichtenstein}\\
  \normalsize
  \includegraphics[width=0.9\linewidth]{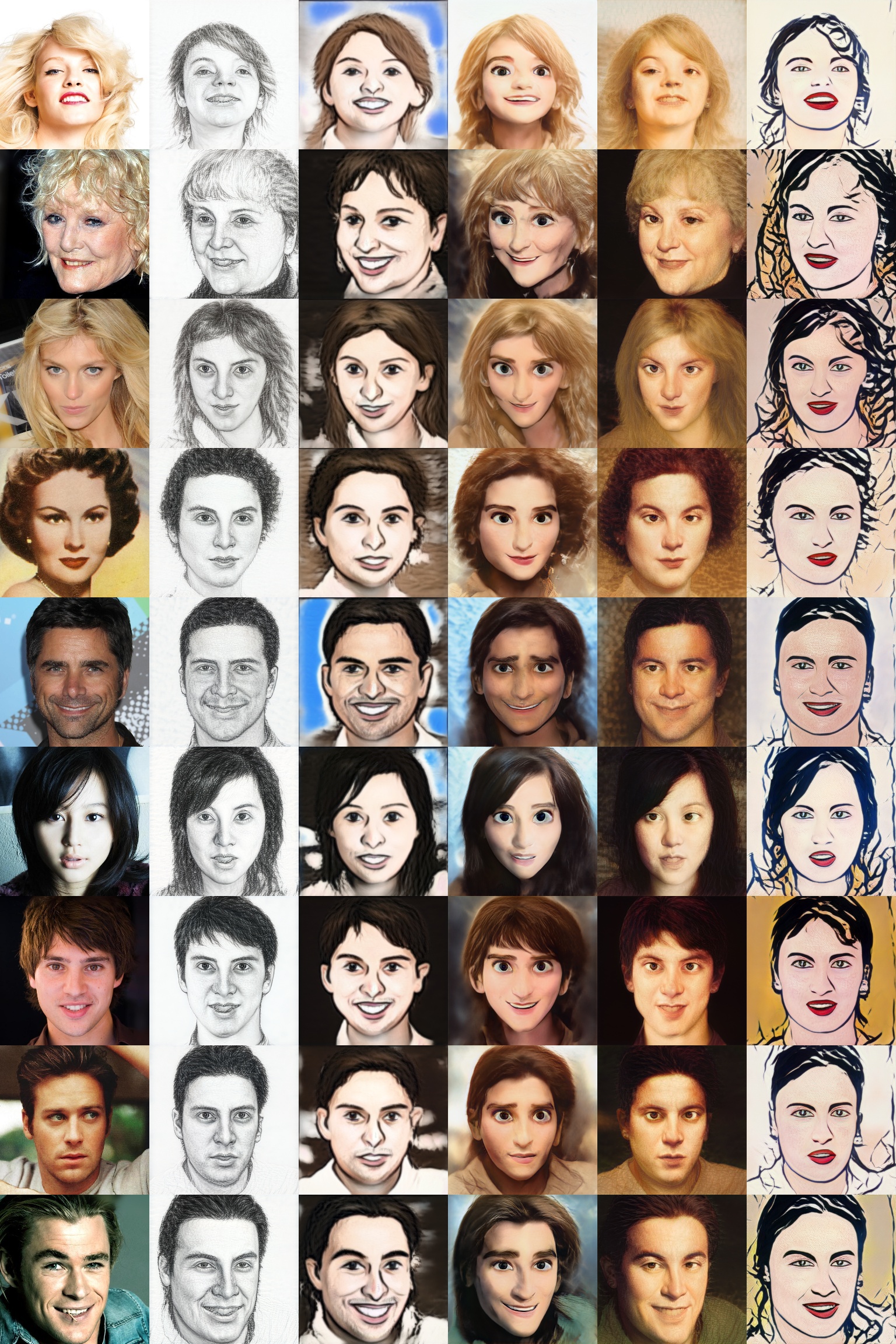}
  \caption{\captionMR{}}
  \label{fig:more_result4}
\end{figure*}

\begin{figure*}
  \centering
  \small
  \makebox[\MRscale\textwidth]{Inputs}
  \makebox[\MRscale\textwidth]{Sketches}
  \makebox[\MRscale\textwidth]{Caricature}
  \makebox[\MRscale\textwidth]{Cartoon}
  \makebox[\MRscale\textwidth]{Raphael}
  \makebox[\MRscale\textwidth]{RoyLichtenstein}\\
  \normalsize
  \includegraphics[width=0.9\linewidth]{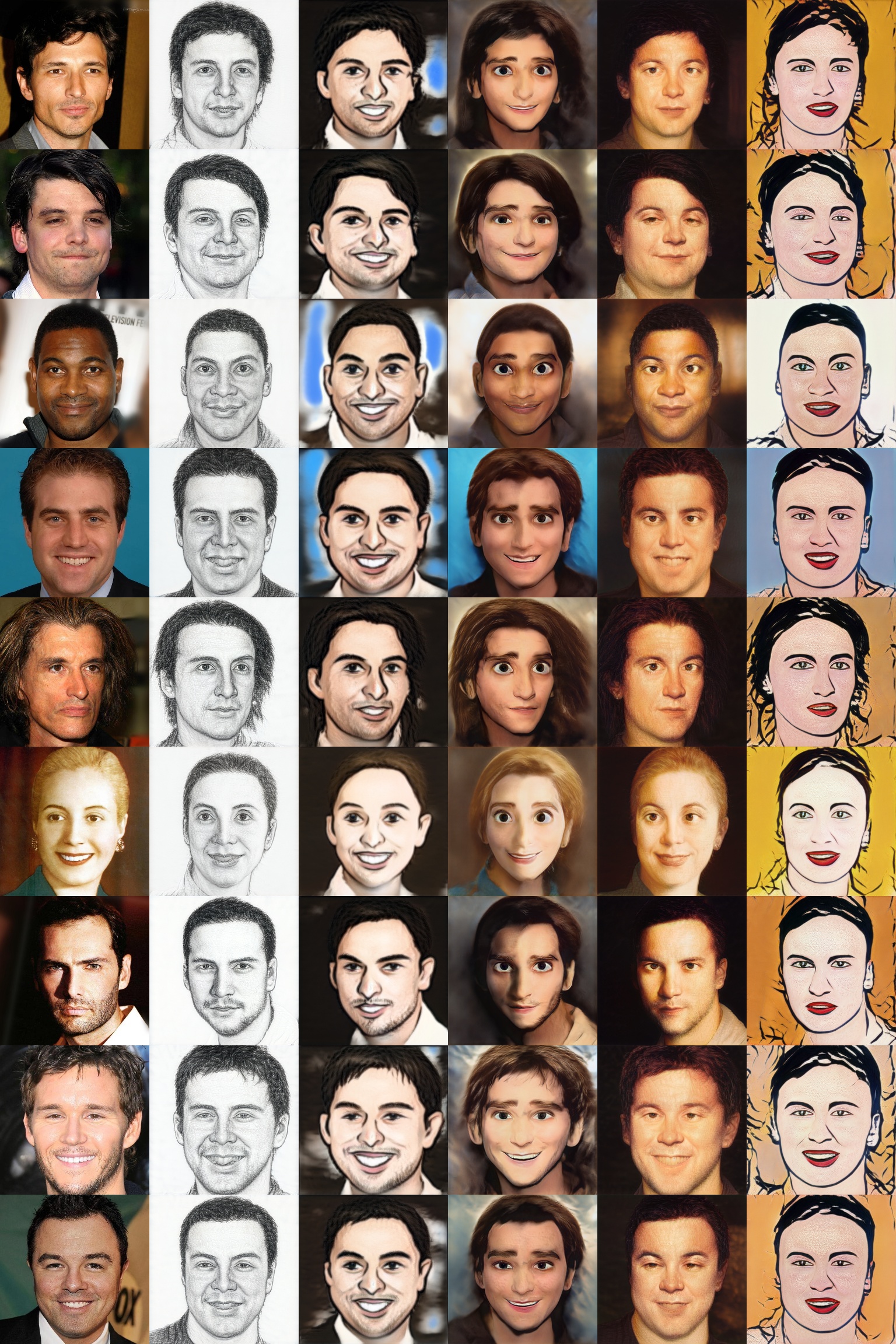}
  \caption{\captionMR{}}
  \label{fig:more_result5}
\end{figure*}

{\bf 10-shot results} are shown in Figs.~\ref{fig:more_result0},~\ref{fig:more_result1},~\ref{fig:more_result2},~\ref{fig:more_result3},~\ref{fig:more_result4},~\ref{fig:more_result5}, including 54 test photos and five different artistic domains,
where the 10 training images of each artistic domain are shown in Fig.~\ref{fig:training}.


\begin{figure}[t]
\centering
\includegraphics[width=.45\linewidth]{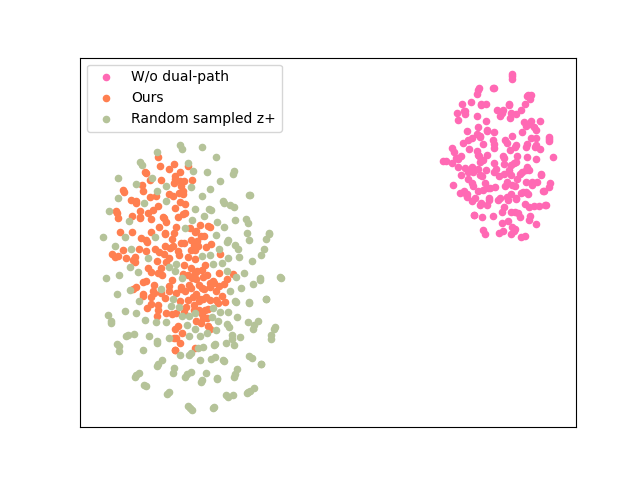}
\vspace{-0.2in}
\caption{t-SNE visualization of dual path training.}
\vspace{-0.1in}
\label{fig:tsne}
\end{figure}



\section{Results on Other Domains}
\label{sec:moredomain}
We test our model on other domains, e.g., Cats and Churches.
The source domain is LSUN~\cite{yu15lsun} Cat or LSUN Church, where we use the StyleGAN2 models pretrained on these datasets.
The target domains include Orange Cat, Blue Cat, Van Gogh House, and Haunted House, where the 10 training images are collected from web and shown in Fig.~\ref{fig:training}.
We directly sample $z+$ latent codes to validate the correctness of our contrastive strategy in these domains.
The testing results are shown in Fig~\ref{fig:cat} and Fig~\ref{fig:church}, our models generate good stylization results and keep the content well.

\begin{figure}[t]
\subfigure[Ablation study on decoder]{
	\begin{minipage}[t]{0.48\linewidth}
        \centering
        \includegraphics[width=1\linewidth]{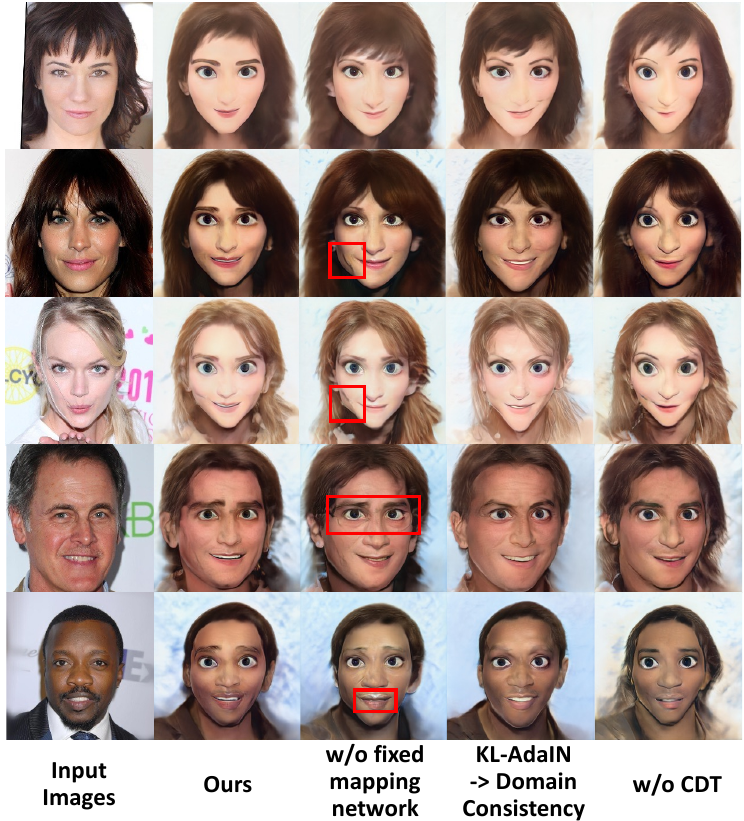}
        \label{fig:more_ablation_decoder}
	\end{minipage}
	}
\subfigure[Analysis of CDT]{
	\begin{minipage}[t]{0.48\linewidth}
        \centering
        \includegraphics[width=1\linewidth]{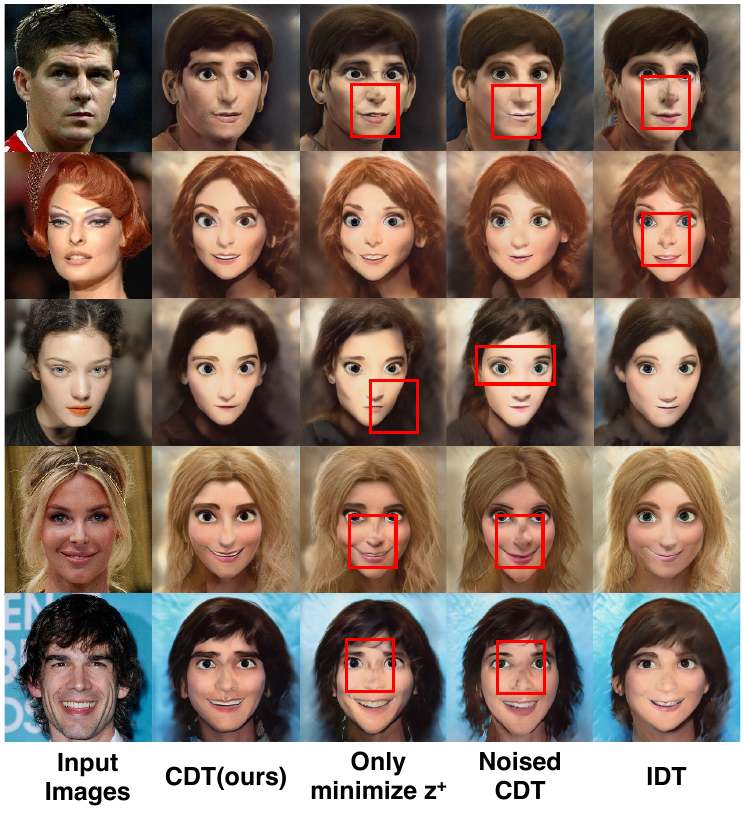}
        \label{fig:analsis-cdt}
	\end{minipage}
	}
\vspace{-0.15in}
\caption{(a) More ablation study results on decoder. The 10 training images are displayed on the left. (b) Analysis of Cross-Domain Triplet loss.}
\vspace{-0.15in}
\end{figure}

\section{More Analysis on Each Component}

\subsection{t-SNE Visualization of the Dual-path Impact}
\label{sec:dual-path}

In this section, we conduct a t-SNE~\cite{van2008visualizing} experiment to visualize the influence of dual-path training strategy used in our encoder.
As shown in Fig.~\ref{fig:tsne}, we compare three groups of $z+$ latent codes:
\begin{enumerate}[nolistsep]
    \item Latent codes generated by our encoder (orange);
    \item Latent codes generated by our encoder without dual-path (pink);
    \item Random sampled latent codes under Gaussian distribution (green).
\end{enumerate}
The results show the dual-path training strategy helps constrain the output latent distribution to follow Gaussian distribution (which is the sampling distribution of decoder input), so that it can better cope with our decoder.


\subsection{Analysis of Sub-encoder}
\label{sec:analysis_sub-encoder}

We compare the reconstruction quality of using linear layers (used in pSp) or attention module, or transformer block as the sub-encoder in our encoder\footnote{Because the feature extractor contains convolution layers and the sub-encoder is after the feature extractor, we don't try convolution layers here.}.
We separately study the three kinds of sub-encoder in $\mathcal W+$ and $\mathcal Z+$ space:
\begin{enumerate}[nolistsep]
    \item Linear layers (fully-connected layers): To explore the effect of linear layers, we try linear settings with one single linear layer and 8 linear layers, and set the input dimension as 512, output dimension as 512, a LeakyReLU following each linear layer;
    \item Attention module: To explore the performance of attention module (we use the implementation in ViT~\cite{dosovitskiy2020image} code), we design the sub-encoder block with 6 attention layers, and set the input dimension as 512, the heads number as 14, the head dimension as 64;
    \item Transformer block: We use ViT~\cite{dosovitskiy2020image} implementation, and set the input dimension as 512, the heads number as 14, the head dimension as 64, the MLP dimension as 1024, and the number of layers as 6.
\end{enumerate}
From Table~\ref{tab:sub-encoder}, we find that transformer block achieves the best reconstruction quality in both kinds of latent spaces, with a closer LPIPS distance than 1 linear layer (-0.02 for $\mathcal Z+$ and -0.02 for $\mathcal W+$), 8 linear layers (-0.03 for $\mathcal Z+$ and -0.09 for $\mathcal W+$) and attention module (-0.25 for $\mathcal Z+$ and -0.13 for $\mathcal W+$).
We also test the FID scores of different encoders on cartoon domain, and results show that transformer block achieves the best stylization.

\begin{table}[t]
	\begin{minipage}[t]{0.5\linewidth}
      \centering
      \caption{Analysis of sub-encoder: average LPIPS distance of reconstruction and cartoon FID score.}
      \label{tab:sub-encoder}
      \scalebox{0.8}{
      \begin{tabular}{l|cc|c}
        \hline
        Models &  \multicolumn{2}{c|}{recon LPIPS $\downarrow$} & FID$\downarrow$ \\
         & $z+$ & $w+$ &  \\
        \hline 
        linear layers (1 layer) & 0.25& 0.21 & 100.18\\
        linear layers (8 layers)& 0.26 & 0.28 & 86.55\\
        attention module (6 layers) & 0.48 &  0.32 & 282.03\\
        transformer block (6 layers)  & {\bf 0.23} & {\bf 0.19} & {\bf 84.93}\\    
        \hline
      \end{tabular}
      }
    \end{minipage}
	\begin{minipage}[t]{0.5\linewidth}
      \centering
      \caption{Ablation study results on decoder.}
      \vspace{+0.15in}
      \label{tab:cartoon ablation}
      \footnotesize
      \scalebox{0.9}{
      \begin{tabular}{l|cc}
        \hline
        Models & FID $\downarrow$ & ld $\downarrow$ \\
        \hline 
        Ours w/o CDT & 113.56& 0.56 \\
        Ours w/o fixed mapping network& 96.87& 0.54\\
        Ours w/o KL-AdaIN& 101.57 & {\bf 0.49} \\
        Ours  & {\bf 84.93} &0.51\\    
        \hline
      \end{tabular}
      }
    \end{minipage}
\end{table}


\subsection{More Ablation Studies on Decoder}
\label{sec:more_ablation}
In the main paper Sec. 4.4, we evaluate the effectness of cross-domain triplet loss in our decoder. In this section we further analyze other components in our decoder.
\begin{itemize}[nolistsep]
    \item KL-AdaIN loss: Apart from CDT loss, we introduce KL-AdaIN loss in our decoder. In this ablation, we replace KL-AdaIN with KL loss (i.e., the cross-domain distance consistency) in~\cite{ojha2021few-shot-gan}. As shown in Fig.~\ref{fig:more_ablation_decoder}(column4), the ablated version has worse style similarity. As shown in Table.~\ref{tab:cartoon ablation}, Ours has much better FID and similar ld with the ablated version.
    \item Fixed mapping network: We found fixing the $Z$-to-$W$ mapping network during adaption helps ease the training. In this ablation, we make the mapping network trainable during adaptation. As shown in Fig.~\ref{fig:more_ablation_decoder}(column3), the ablated version contains more artifacts than ours. As shown in Table.~\ref{tab:cartoon ablation}, Ours outperforms the ablated version on both metrics.
\end{itemize}

\subsection{Detailed Analysis on Triplet Loss}
\label{sec:triplet}
In the main paper Sec. 3.1, we describe our Cross-Domain Triplet loss (CDT).
\begin{equation}
  \begin{aligned}
  \mathcal{L}_{cdt} = \mathbb{E}_{\{z_i \sim p_z(z)\}}\max({d^+}(z_i) - {d^-}(z_i) + \alpha, 0)
  \label{eq:cdt_loss3}
  \end{aligned}
\end{equation}
\begin{equation}
  \begin{aligned}
  {d^+}(z_i)=\mathcal{L}_{d}(\mathcal{G}_{s}(z_i),\mathcal{G}_{t}(z_i)) 
  \label{eq:cdt_loss4}
  \end{aligned}
\end{equation}
\begin{equation}
  \begin{aligned}
  {d^-}(z_i)=\frac{1}{m-1}\sum_{j, j\neq i}^m \mathcal{L}_{d}(\mathcal{G}_{s}(z_i),\mathcal{G}_{t}(z_j)),
  \label{eq:cdt_loss5}
  \end{aligned}
\end{equation}
And In the main paper Sec. 4.5 and Table 5, we validate the the design of cross-domain triplet loss with three different designs.

In this section, we describe the three comparison designs in detail, and provide more qualitative comparisons (Fig.~\ref{fig:analsis-cdt}):
\begin{enumerate}[nolistsep]
\item
{\bf Only minimizing $d+$.}
In this ablation, given two sampled latent codes ${z_i}$, we directly minimize the LPIPS distance between the source domain image and target domain image ($d+$). Fig.~\ref{fig:analsis-cdt}(column3) shows this hurts the stylization, while our CDT ($d-$ counteracts the style difference in $d+$) achieves better stylization.
\item {\bf In-Domain Triplet loss (IDT).}
In this ablation, we only compare generated results within the target domain, instead of between source and target domains.
Given two sampled latent codes ${z_i}$ and ${z_j}$, we sample a close ${z_i}^* = {z_i} + \Delta{z_i}, \Delta{z_i} \sim \mathcal{N}(0,0.1)$. We set the anchor as $\mathcal{G}_t(z_i)$, the positive example as $\mathcal{G}_t({z_i}^*)$, and the negative example as $\mathcal{G}_t(z_j)$, where $\mathcal{G}_t$ is the target decoder. Fig.~\ref{fig:analsis-cdt}(column5) shows its results contain artifacts, while our CDT (cross-domain distance) achieves better results.
\item {\bf Noised Cross-Domain Triplet loss (Noised CDT).}
In this ablation, we change the positive example from $\mathcal{G}_t(z_i)$ to $\mathcal{G}_t(z_i^*)$, where ${z_i}^*$ is a close latent code to $z_i$, ${z_i}^* = {z_i} + \Delta{z_i}, \Delta{z_i} \sim \mathcal{N}(0,0.1)$.
The anchor is $\mathcal{G}_s(z_i)$, the positive example is $\mathcal{G}_t({z_i}^*)$, and the negative example is $\mathcal{G}_t(z_j)$. Fig.~\ref{fig:analsis-cdt}(column4) shows that the results are worse in keeping identity, while our CDT (same $z_i$) achieves better results.
\end{enumerate}


\section{Detailed Network Architecture and Hyper-Parameters}
\label{sec:network_detail}

\subsection{Few-shot Domain Adaptation Decoder}
{\bf Architecture: }
We adopt StyleGAN2 architecture~\cite{Karras2019stylegan2} for our decoder, to map $\mathcal{Z}+$ space latent codes into artistic portraits.
We use adversarial training for adapting the decoder, and use two discriminators, an
image discriminator, a patch-level discriminator
following the implementation from Few-shot-GAN-adaptation~\cite{ojha2021few-shot-gan}.

{\bf Training details and hyper-parameters:}
We adopt a pretrained StyleGAN2 on FFHQ as the base model and then adapt the base model to our target artistic domain.

For 10-shot training,
we set $\lambda_{adv} = 1.0$, $\lambda_{kladain} = 1000$ in main paper Eq.(8) for all target artistic domains, and set different $\lambda_{cdt}$ for different artistic domains as follows:
\begin{enumerate}[nolistsep]
    \item $\lambda_{cdt} = 0.05$ for Sketches and Raphael domains;
    \item $\lambda_{cdt} = 0.02$ for Caricature domain, Cat domain and Church domain tasks;
    \item $\lambda_{cdt} = 0.005$ for Cartoon, Roy Lichtenstein and Sunglasses domains.
\end{enumerate}
We train 5000 iterations for Sketches domain, 3000 iterations for Raphael domain and Caricature domains, 2000 iterations for Sunglasses domain, 1250 iterations for Roy Lichtenstein domain, and 1000 iterations for Cartoon domain.
We set learning rate $lr = 0.002$ for all of the above decoder training. 

For 1-shot training, we set $\lambda_{cdt} = 0.005$ for face domain tasks, and train about 600 iterations for all the target domains.

For Cross-Domain Triplet Loss calculation, we find that adding a weight for $d^+$ in main paper Eq.(2) helps reduce artifacts, i.e., 
\begin{equation}
\mathcal{L}_{cdt} = \mathbb{E}_{\{z_i \sim p_z(z)\}}\max(w\cdot {d^+}(z_i) - {d^-}(z_i) + \alpha, 0)
\end{equation}
We set $w=1.5$ for Sketches domain, $w=2.0$ for other domains, and $\alpha = 2$ for all target artistic domains.

\subsection{Style Encoder}
\label{STE}
{\bf Architecture: }Our encoder consists of a feature extractor and a sub-encoder.
For the feature extractor, we follow pSp~\cite{richardson2021encoding} and adopt its FPN~\cite{Lin2017FPN} based feature extractor design.
For the sub-encoder, we compared different designs (detailed in Sec.~\ref{sec:analysis_sub-encoder}) and found transformer-based architecture achieves the best results.

{\bf Training details: }
We first train Style Encoder from scratch on FFHQ~\cite{stylegan} dataset for 170,000 iterations in path-1 (mentioned in main paper section 3.2), and use the model as pretrained encoder model.
Then, we train Style Encoder in dual path setting (both path-1 and path-2) for 70,000 iterations.

{\bf Training details for encoder ablation studies: }
For ablation study on dual-path, we train another 70,000 iterations only in path-1 from the pretrained encoder model to match our full encoder (dual path) setting.

{\bf Hyper-parameters: }
In the above training process, we set $\lambda_{L_2} = 1.0$, $\lambda_{lpips} =0.8 $, $\lambda_{reg}=0$, $\lambda_{iden} = 0.1$ in main paper Eq.(9).
And set $\lambda_{z-predict}=0.1$ and $\lambda_{path1} = 1$ in main paper Eq.(11).
We set the learning rate as $lr=0.0001$.

{\bf For more details, we provide the source code for closer inspection.}


\section{Comparison with Neural Style Transfer and Image-to-Image Translation Methods}
\label{sec:campare}
\begin{figure}[t]
  \centering
  \includegraphics[width=0.60\linewidth]{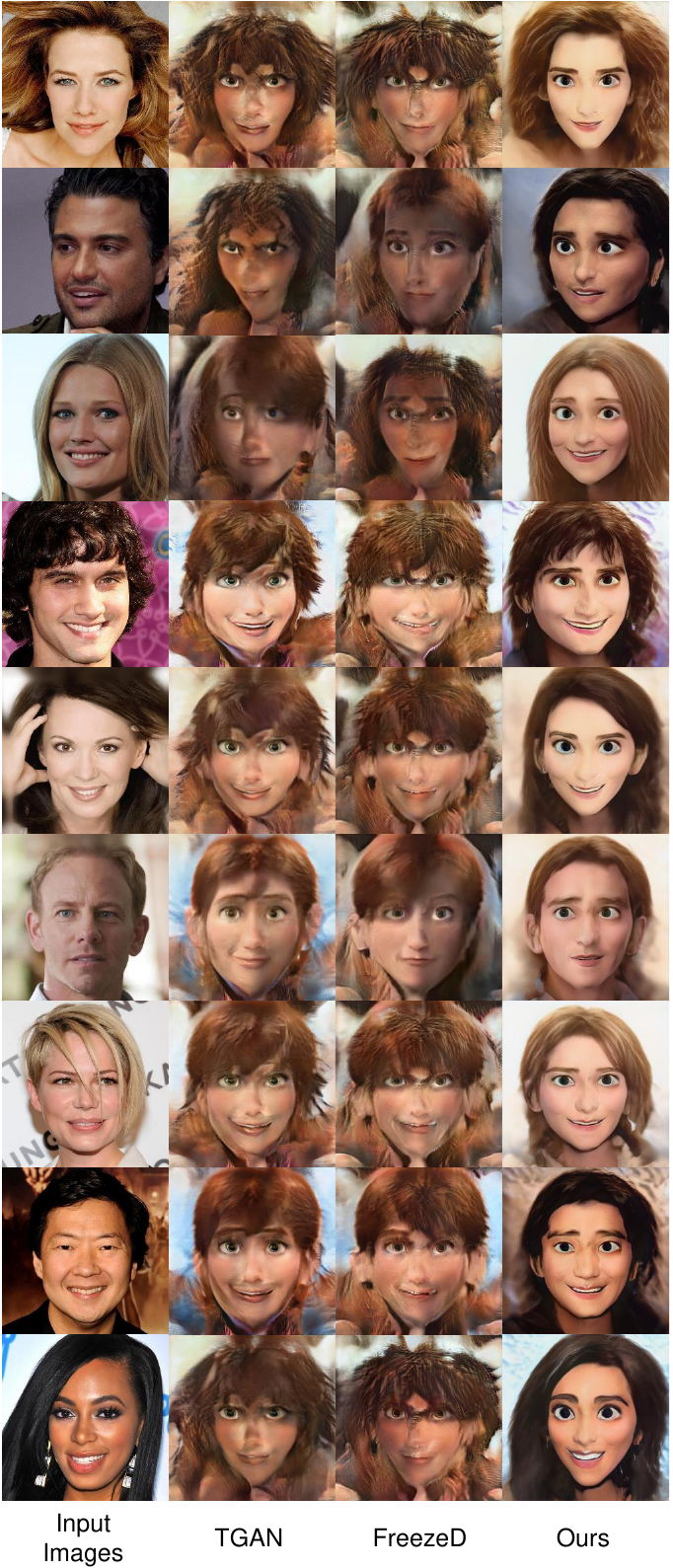}
  \caption{More comparisons with TGAN~\cite{wang2018transferring} and FreezeD~\cite{mo2020freeze}}.
  \label{fig:compare_more1}
\end{figure}

In this section, we provide more comparisons.
For TGAN~\cite{wang2018transferring} and FreezeD~\cite{mo2020freeze}, in the main paper we use an anchor sample space~\cite{ojha2021few-shot-gan} for $z$ to prevent overfitting. Here, we use the full sample space (i.e. sample $z$ from Gaussian distribution), to evaluate the two comparison methods in their original settings.
As shown in Fig.~\ref{fig:compare_more1}, the results of these two methods are prone to overfitting.

\begin{figure*}
  \centering
  \includegraphics[width=1\linewidth]{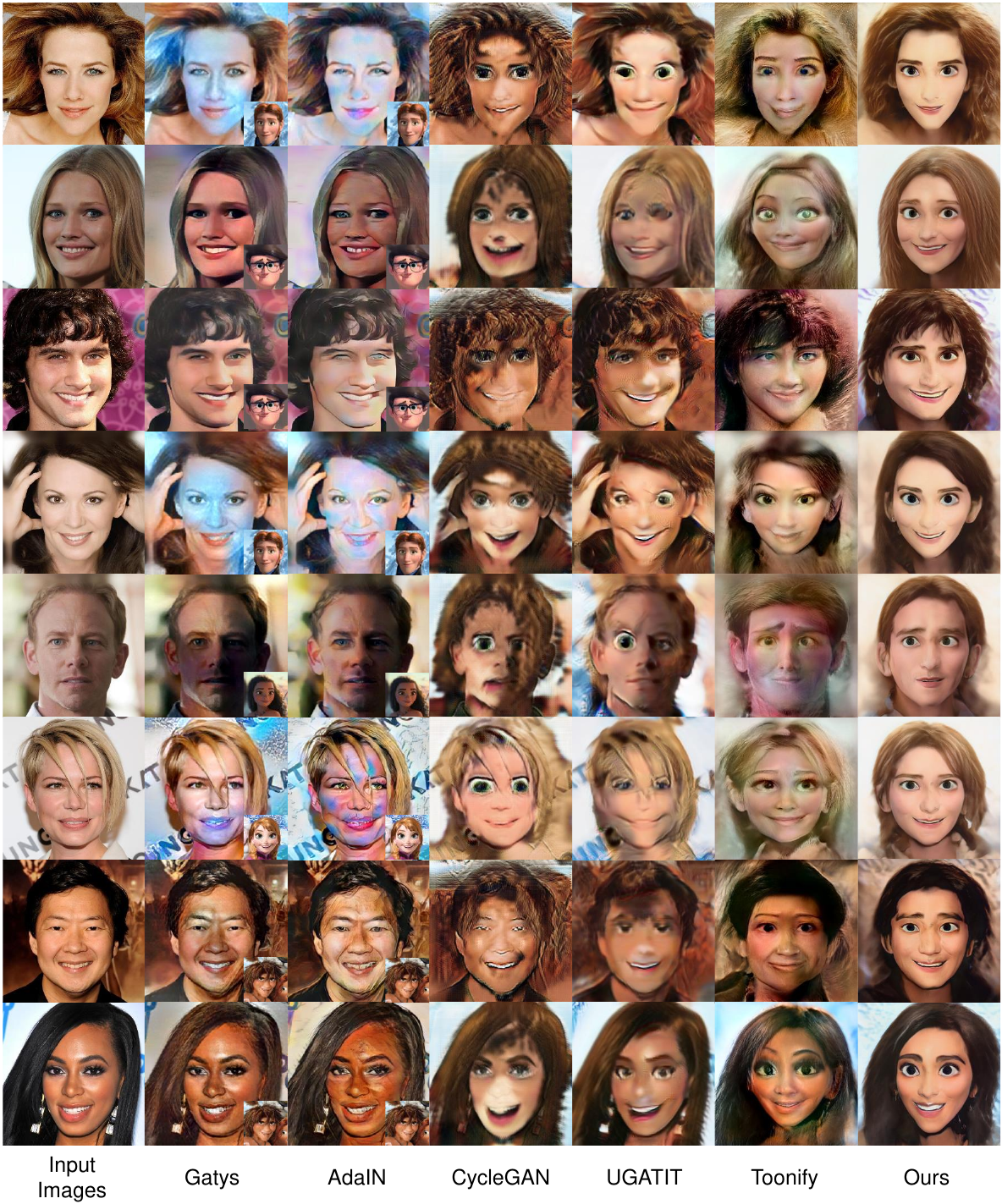}
  \vspace{-0.1in}
  \caption{Comparisons with more neural style transfer and image-to-image translation methods: Gatys~\cite{GatysEB16}, AdaIN~\cite{huang2017arbitrary}, CycleGAN~\cite{cyclegan}, UGATIT~\cite{Kim2020U-GAT-IT:}, and  Toonify~\cite{pinkney2020resolution}}.
  \vspace{-0.1in}
  \label{fig:compare_more2}
\end{figure*}

We further compare with 5 neural style transfer, image-to-image translation methods under 10-shot setting:
Gatys\cite{GatysEB16},
AdaIN\cite{huang2017arbitrary},
CycleGAN\cite{cyclegan},
UGATIT~\cite{Kim2020U-GAT-IT:},
and Toonify\cite{pinkney2020resolution}.
Qualitative comparison results are shown in Fig.~\ref{fig:compare_more2}.
We find neural style transfer methods (Gatys, AdaIN) sometimes fail to capture the target cartoon style and generate results with artifacts.
CycleGAN and UGATIT results are of lower quality under few-shot setting. Toonify results also contain artifacts.
In comparison, our method generates high-quality artistic portraits.


\end{subappendices}

%
%
\bibliographystyle{splncs04}
\bibliography{egbib}

\begin{thebibliography}{10}
\providecommand{\url}[1]{\texttt{#1}}
\providecommand{\urlprefix}{URL }
\providecommand{\doi}[1]{https://doi.org/#1}

\bibitem{image2stylegan}
Abdal, R., Qin, Y., Wonka, P.: Image2stylegan: How to embed images into the
  stylegan latent space? In: 2019 IEEE/CVF International Conference on Computer
  Vision (ICCV). pp. 4431--4440 (2019)

\bibitem{image2stylegan++}
Abdal, R., Qin, Y., Wonka, P.: Image2stylegan++: How to edit the embedded
  images? In: 2020 IEEE/CVF Conference on Computer Vision and Pattern
  Recognition (CVPR). pp. 8293--8302 (2020)

\bibitem{abdal2021styleflow}
Abdal, R., Zhu, P., Mitra, N.J., Wonka, P.: Styleflow: Attribute-conditioned
  exploration of stylegan-generated images using conditional continuous
  normalizing flows. ACM Transactions on Graphics (TOG)  \textbf{40}(3),  1--21
  (2021)

\bibitem{alaluf2021restyle}
Alaluf, Y., Patashnik, O., Cohen-Or, D.: Restyle: A residual-based stylegan
  encoder via iterative refinement. In: Proceedings of the IEEE/CVF
  International Conference on Computer Vision (ICCV) (2021)

\bibitem{carion2020end}
Carion, N., Massa, F., Synnaeve, G., Usunier, N., Kirillov, A., Zagoruyko, S.:
  End-to-end object detection with transformers. In: European conference on
  computer vision. pp. 213--229. Springer (2020)

\bibitem{ChoiCKH0C18}
Choi, Y., Choi, M., Kim, M., Ha, J., Kim, S., Choo, J.: Star{GAN}: Unified
  generative adversarial networks for multi-domain image-to-image translation.
  In: {IEEE} Conference on Computer Vision and Pattern Recognition ({CVPR}).
  pp. 8789--8797 (2018)

\bibitem{chong2021jojogan}
Chong, M.J., Forsyth, D.: Jojogan: One shot face stylization. arXiv preprint
  arXiv:2112.11641  (2021)

\bibitem{deng2019arcface}
Deng, J., Guo, J., Xue, N., Zafeiriou, S.: Arcface: Additive angular margin
  loss for deep face recognition. In: Proceedings of the IEEE/CVF Conference on
  Computer Vision and Pattern Recognition (CVPR). pp. 4690--4699 (2019)

\bibitem{dosovitskiy2020image}
Dosovitskiy, A., Beyer, L., Kolesnikov, A., Weissenborn, D., Zhai, X.,
  Unterthiner, T., Dehghani, M., Minderer, M., Heigold, G., Gelly, S., et~al.:
  An image is worth 16x16 words: Transformers for image recognition at scale.
  arXiv preprint arXiv:2010.11929  (2020)

\bibitem{gal2021stylegan}
Gal, R., Patashnik, O., Maron, H., Chechik, G., Cohen-Or, D.: Stylegan-nada:
  Clip-guided domain adaptation of image generators. arXiv preprint
  arXiv:2108.00946  (2021)

\bibitem{GatysEB16}
Gatys, L.A., Ecker, A.S., Bethge, M.: Image style transfer using convolutional
  neural networks. In: {IEEE} Conference on Computer Vision and Pattern
  Recognition ({CVPR}). pp. 2414--2423 (2016)

\bibitem{girshick2015fast}
Girshick, R.: Fast r-cnn. In: Proceedings of the IEEE international conference
  on computer vision. pp. 1440--1448 (2015)

\bibitem{GoodfellowPMXWOCB14}
Goodfellow, I.J., Pouget{-}Abadie, J., Mirza, M., Xu, B., Warde{-}Farley, D.,
  Ozair, S., Courville, A.C., Bengio, Y.: Generative adversarial nets. In:
  Advances in Neural Information Processing Systems ({NeurIPS}). pp. 2672--2680
  (2014)

\bibitem{heusel2017gans}
Heusel, M., Ramsauer, H., Unterthiner, T., Nessler, B., Hochreiter, S.: Gans
  trained by a two time-scale update rule converge to a local nash equilibrium.
  Advances in neural information processing systems  \textbf{30} (2017)

\bibitem{huang2017arbitrary}
Huang, X., Belongie, S.: Arbitrary style transfer in real-time with adaptive
  instance normalization. In: Proceedings of the IEEE International Conference
  on Computer Vision. pp. 1501--1510 (2017)

\bibitem{HuangLBK18}
Huang, X., Liu, M., Belongie, S.J., Kautz, J.: Multimodal unsupervised
  image-to-image translation. In: 15th European Conference ({ECCV}). pp.
  179--196 (2018)

\bibitem{pix2pix}
Isola, P., Zhu, J.Y., Zhou, T., Efros, A.A.: Image-to-image translation with
  conditional adversarial networks. In: 2017 IEEE Conference on Computer Vision
  and Pattern Recognition (CVPR). pp. 5967--5976 (2017)

\bibitem{jang2021stylecarigan}
Jang, W., Ju, G., Jung, Y., Yang, J., Tong, X., Lee, S.: Stylecarigan:
  caricature generation via stylegan feature map modulation. ACM Transactions
  on Graphics (TOG)  \textbf{40}(4),  1--16 (2021)

\bibitem{JohnsonAF16}
Johnson, J., Alahi, A., Fei{-}Fei, L.: Perceptual losses for real-time style
  transfer and super-resolution. In: 14th European Conference ({ECCV}). pp.
  694--711 (2016)

\bibitem{karras2017progressive}
Karras, T., Aila, T., Laine, S., Lehtinen, J.: Progressive growing of gans for
  improved quality, stability, and variation. arXiv preprint arXiv:1710.10196
  (2017)

\bibitem{Karras2020ada}
Karras, T., Aittala, M., Hellsten, J., Laine, S., Lehtinen, J., Aila, T.:
  Training generative adversarial networks with limited data. In: Proc. NeurIPS
  (2020)

\bibitem{karras2021alias}
Karras, T., Aittala, M., Laine, S., H{\"a}rk{\"o}nen, E., Hellsten, J.,
  Lehtinen, J., Aila, T.: Alias-free generative adversarial networks. arXiv
  preprint arXiv:2106.12423  (2021)

\bibitem{stylegan}
Karras, T., Laine, S., Aila, T.: A style-based generator architecture for
  generative adversarial networks. In: IEEE/CVF Conference on Computer Vision
  and Pattern Recognition (CVPR). pp. 4396--4405 (2019)

\bibitem{Karras2019stylegan2}
Karras, T., Laine, S., Aittala, M., Hellsten, J., Lehtinen, J., Aila, T.:
  Analyzing and improving the image quality of {StyleGAN}. In: IEEE/CVF
  Conference on Computer Vision and Pattern Recognition (CVPR) (2020)

\bibitem{Kim2020U-GAT-IT:}
Kim, J., Kim, M., Kang, H., Lee, K.H.: U-gat-it: Unsupervised generative
  attentional networks with adaptive layer-instance normalization for
  image-to-image translation. In: International Conference on Learning
  Representations (2020), \url{https://openreview.net/forum?id=BJlZ5ySKPH}

\bibitem{li2020ewc}
Li, Y., Zhang, R., Lu, J., Shechtman, E.: Few-shot image generation with
  elastic weight consolidation. In: Advances in Neural Information Processing
  Systems (2020)

\bibitem{Lin2017FPN}
Lin, T.Y., Dollár, P., Girshick, R., He, K., Hariharan, B., Belongie, S.:
  Feature pyramid networks for object detection. In: 2017 IEEE Conference on
  Computer Vision and Pattern Recognition (CVPR). pp. 936--944 (2017).
  \doi{10.1109/CVPR.2017.106}

\bibitem{LiuBK17}
Liu, M., Breuel, T., Kautz, J.: Unsupervised image-to-image translation
  networks. In: Advances in Neural Information Processing Systems ({NeurIPS}).
  pp. 700--708 (2017)

\bibitem{liu2019few}
Liu, M.Y., Huang, X., Mallya, A., Karras, T., Aila, T., Lehtinen, J., Kautz,
  J.: Few-shot unsupervised image-to-image translation. In: Proceedings of the
  IEEE/CVF International Conference on Computer Vision. pp. 10551--10560 (2019)

\bibitem{van2008visualizing}
Van~der Maaten, L., Hinton, G.: Visualizing data using t-sne. Journal of
  machine learning research  \textbf{9}(11) (2008)

\bibitem{mo2020freeze}
Mo, S., Cho, M., Shin, J.: Freeze the discriminator: a simple baseline for
  fine-tuning gans. arXiv preprint arXiv:2002.10964  (2020)

\bibitem{noguchi2019image}
Noguchi, A., Harada, T.: Image generation from small datasets via batch
  statistics adaptation. In: Proceedings of the IEEE/CVF International
  Conference on Computer Vision. pp. 2750--2758 (2019)

\bibitem{ojha2021few-shot-gan}
Ojha, U., Li, Y., Lu, C., Efros, A.A., Lee, Y.J., Shechtman, E., Zhang, R.:
  Few-shot image generation via cross-domain correspondence. In: CVPR (2021)

\bibitem{park2020contrastive}
Park, T., Efros, A.A., Zhang, R., Zhu, J.Y.: Contrastive learning for unpaired
  image-to-image translation. In: European Conference on Computer Vision. pp.
  319--345 (2020)

\bibitem{pinkney2020resolution}
Pinkney, J.N., Adler, D.: Resolution dependent gan interpolation for
  controllable image synthesis between domains. arXiv preprint arXiv:2010.05334
   (2020)

\bibitem{clip}
Radford, A., Kim, J.W., Hallacy, C., Ramesh, A., Goh, G., Agarwal, S., Sastry,
  G., Askell, A., Mishkin, P., Clark, J., et~al.: Learning transferable visual
  models from natural language supervision. In: International Conference on
  Machine Learning. pp. 8748--8763. PMLR (2021)

\bibitem{richardson2021encoding}
Richardson, E., Alaluf, Y., Patashnik, O., Nitzan, Y., Azar, Y., Shapiro, S.,
  Cohen-Or, D.: Encoding in style: a stylegan encoder for image-to-image
  translation. In: 2021 IEEE/CVF Conference on Computer Vision and Pattern
  Recognition (CVPR) (June 2021)

\bibitem{schroff2015facenet}
Schroff, F., Kalenichenko, D., Philbin, J.: Facenet: A unified embedding for
  face recognition and clustering. In: Proceedings of the IEEE conference on
  computer vision and pattern recognition. pp. 815--823 (2015)

\bibitem{simonyan2014very}
Simonyan, K., Zisserman, A.: Very deep convolutional networks for large-scale
  image recognition. arXiv preprint arXiv:1409.1556  (2014)

\bibitem{agilegan}
Song, G., Luo, L., Liu, J., Ma, W.C., Lai, C., Zheng, C., Cham, T.J.: Agilegan:
  Stylizing portraits by inversion-consistent transfer learning. ACM Trans.
  Graph.  \textbf{40}(4) (Jul 2021). \doi{10.1145/3450626.3459771},
  \url{https://doi.org/10.1145/3450626.3459771}

\bibitem{sun2021rethinking}
Sun, Z., Cao, S., Yang, Y., Kitani, K.M.: Rethinking transformer-based set
  prediction for object detection. In: Proceedings of the IEEE/CVF
  International Conference on Computer Vision. pp. 3611--3620 (2021)

\bibitem{tewari2020stylerig}
Tewari, A., Elgharib, M., Bharaj, G., Bernard, F., Seidel, H.P., P{\'e}rez, P.,
  Zollhofer, M., Theobalt, C.: Stylerig: Rigging stylegan for 3d control over
  portrait images. In: Proceedings of the IEEE/CVF Conference on Computer
  Vision and Pattern Recognition. pp. 6142--6151 (2020)

\bibitem{e4e}
Tov, O., Alaluf, Y., Nitzan, Y., Patashnik, O., Cohen-Or, D.: Designing an
  encoder for stylegan image manipulation. ACM Trans. Graph.  \textbf{40}(4)
  (Jul 2021). \doi{10.1145/3450626.3459838},
  \url{https://doi.org/10.1145/3450626.3459838}

\bibitem{Wang0ZTKC18}
Wang, T., Liu, M., Zhu, J., Tao, A., Kautz, J., Catanzaro, B.: High-resolution
  image synthesis and semantic manipulation with conditional {GAN}s. In: {IEEE}
  Conference on Computer Vision and Pattern Recognition ({CVPR}). pp.
  8798--8807 (2018)

\bibitem{wang2008face}
Wang, X., Tang, X.: Face photo-sketch synthesis and recognition. IEEE
  transactions on pattern analysis and machine intelligence  \textbf{31}(11),
  1955--1967 (2008)

\bibitem{wang2020minegan}
Wang, Y., Gonzalez-Garcia, A., Berga, D., Herranz, L., Khan, F.S., Weijer,
  J.v.d.: Minegan: effective knowledge transfer from gans to target domains
  with few images. In: Proceedings of the IEEE/CVF Conference on Computer
  Vision and Pattern Recognition. pp. 9332--9341 (2020)

\bibitem{wang2018transferring}
Wang, Y., Wu, C., Herranz, L., van~de Weijer, J., Gonzalez-Garcia, A.,
  Raducanu, B.: Transferring gans: generating images from limited data. In:
  Proceedings of the European Conference on Computer Vision (ECCV). pp.
  218--234 (2018)

\bibitem{YangRX021}
Yang, T., Ren, P., Xie, X., Zhang, L.: {GAN} prior embedded network for blind
  face restoration in the wild. In: {IEEE} Conference on Computer Vision and
  Pattern Recognition, {CVPR}. pp. 672--681 (2021)

\bibitem{yaniv2019face}
Yaniv, J., Newman, Y., Shamir, A.: The face of art: landmark detection and
  geometric style in portraits. ACM Transactions on graphics (TOG)
  \textbf{38}(4),  1--15 (2019)

\bibitem{YiLLR19}
Yi, R., Liu, Y., Lai, Y., Rosin, P.L.: Apdrawing{GAN}: Generating artistic
  portrait drawings from face photos with hierarchical {GAN}s. In: {IEEE}
  Conference on Computer Vision and Pattern Recognition ({CVPR}). pp.
  10743--10752 (2019)

\bibitem{YiZTG17}
Yi, Z., Zhang, H.R., Tan, P., Gong, M.: Dualgan: Unsupervised dual learning for
  image-to-image translation. In: {IEEE} International Conference on Computer
  Vision ({ICCV}). pp. 2868--2876 (2017)

\bibitem{yu15lsun}
Yu, F., Zhang, Y., Song, S., Seff, A., Xiao, J.: Lsun: Construction of a
  large-scale image dataset using deep learning with humans in the loop. arXiv
  preprint arXiv:1506.03365  (2015)

\bibitem{zhang2018unreasonable}
Zhang, R., Isola, P., Efros, A.A., Shechtman, E., Wang, O.: The unreasonable
  effectiveness of deep features as a perceptual metric. In: Proceedings of the
  IEEE conference on computer vision and pattern recognition. pp. 586--595
  (2018)

\bibitem{cyclegan}
Zhu, J.Y., Park, T., Isola, P., Efros, A.A.: Unpaired image-to-image
  translation using cycle-consistent adversarial networks. In: 2017 IEEE
  International Conference on Computer Vision (ICCV). pp. 2242--2251 (2017)

\bibitem{zhu2021mind}
Zhu, P., Abdal, R., Femiani, J., Wonka, P.: Mind the gap: Domain gap control
  for single shot domain adaptation for generative adversarial networks. arXiv
  preprint arXiv:2110.08398  (2021)

\end{thebibliography}

\end{document}